\documentclass[journal]{IEEEtran}
\usepackage{lineno,hyperref}
\usepackage{bm}
\usepackage{mathrsfs}
\usepackage{amsmath,cases}
\usepackage{amssymb}
\usepackage{textcomp}
\usepackage{graphicx}
\usepackage{subfigure}
\usepackage{epstopdf}
\usepackage{xcolor}
\usepackage{overpic}
\modulolinenumbers[5]

\ifCLASSINFOpdf

\else

\fi

\begin{document}

\title{Study of Robust Distributed Diffusion RLS Algorithms with Side Information for Adaptive Networks}

\author{Yi~Yu,
        ~Haiquan Zhao,~\IEEEmembership{Senior Member,~IEEE},
        ~Rodrigo C.~de Lamare,~\IEEEmembership{Senior Member,~IEEE},
        ~Yuriy Zakharov,~\IEEEmembership{Senior Member,~IEEE},
        ~and
        Lu~Lu
\thanks{This work is partially supported by National Nature Science Foundation of P.R. China (Nos: 61871461, 61571374, 61433011, 11472297). The work of Y. Zakharov is partly supported by the UK Engineering and Physical Sciences Research Council (EPSRC) through Grant EP/R003297/1. The work of L. Lu is supported by China Postdoctoral Science Foundation Funded Project under Grant 2018M640916. An earlier version of this work was reported in the conference presentation \emph{IEEE International Conf. Acoustics, Speech and Signal Processing (ICASSP)}, Alberta, Canada, April 2018 \cite{YYu2018}.}

\thanks{Y. Yu is with the School of Information Engineering, Southwest University of Science and Technology, Mianyang, 621010, China (e-mail: yuyi\_xyuan@163.com)}
\thanks{H. Zhao is with the School of Electrical Engineering, Southwest Jiaotong University, Chengdu, 610031, China. (e-mail: hqzhao\_swjtu@126.com).}
\thanks{R. C. de Lamare is with the CETUC, PUC-Rio, Rio de Janeiro 22451-900, Brazil, and also with the Department of Electronic Engineering, University of York, York YO10 5DD, U.K. (e-mail: rcdl500@ohm.york.ac.uk).}
\thanks{Y. Zakharov is with the Department of Electronic Engineering, University of York, York YO10 5DD, U.K. (e-mail: yury.zakharov@york.ac.uk).}
\thanks{L. Lu is with the School of Electronics and Information Engineering, Sichuan University, Chengdu, China. (lulu19900303@126.com).}
}

\markboth{}%
{Shell \MakeLowercase{\textit{et al.}}: Bare Demo of IEEEtran.cls for IEEE Journals}

\maketitle

\begin{abstract}
This work develops robust diffusion recursive least squares algorithms to mitigate the performance degradation often experienced in networks of agents in the presence of impulsive noise. The first algorithm minimizes an exponentially weighted least-squares cost function subject to a time-dependent constraint on the squared norm of the intermediate update at each node. A recursive strategy for computing the constraint is proposed using side information from the neighboring nodes to further improve the robustness. We also analyze the mean-square convergence behavior of the proposed algorithm. The second proposed algorithm is a modification of the first one based on the dichotomous coordinate descent iterations. It has a performance similar to that of the former, however its complexity is significantly lower especially when input regressors of agents have a shift structure and it is well suited to practical implementation. Simulations show the superiority of the proposed algorithms over previously reported techniques in various impulsive noise scenarios.
\end{abstract}

\begin{IEEEkeywords}
Distributed algorithms, diffusion cooperation, dichotomous coordinate-descent, impulsive noises, recursive least squares algorithms.
\end{IEEEkeywords}

\IEEEpeerreviewmaketitle

\section{Introduction}
\IEEEPARstart{O}{ver} the past decade, distributed parameter estimation over wireless sensor networks with multiple nodes (agents) has attracted much attention. It only relies on the local data exchange between interconnected nodes, and therefore removes the requirement of a powerful central processor and, as such, reduces communications bandwidth of the traditional centralized estimation whilst retaining similar estimation performance~\cite{sayed2014adaptive, sayed2014adaptation}. Distributed estimation has been applied to target localization \cite{tu2011mobile}, clustering \cite{sayed2013diffusion}, frequency estimation \cite{kanna2015distributed} and spectrum estimation in Cognitive radio (CR) \cite{di2013distributed, miller2016distributed}.
\subsection{Prior and Related Work}
According to the cooperation strategies between interconnected nodes, existing algorithms can be categorized as incremental \cite{li2010distributed}, consensus \cite{tu2012diffusion}, and diffusion~\cite{lopes2008diffusion, chen2012diffusion,xu2015adaptive,LU2018243} types. Among these, the diffusion strategy is popular, because it does not require a Hamiltonian cycle path as in the incremental type, thereby it is more robust to nodes/links failures; it is stable and shows a faster convergence rate and a lower mean-square error than that of the consensus approach. Several diffusion algorithms were proposed, e.g., diffusion least mean square (dLMS) algorithm~\cite{lopes2008diffusion} and its variable step size variants~\cite{lee2015variable,han2017non}.

In practice, the measurements can be corrupted by non-Gaussian noise
with impulsiveness. Impulsive noise has small occurrence probability
but much higher amplitude than the nominal measurements. It may
occur due to atmospheric phenomena, or man-made due to either
electric machinery in the operation
environment~\cite{aifir,miller1976detection,blackard1993measurements,jio,rrsgp,spa,zoubir2012robust,smtvb,jidf,jidf_echo,sjidf,ccg,jiocdma,jiomimo,tds,mbdf,rrstap,l1stap,l1stap2,rccm,dfjio,locsme,rrser,rdrcb,rrdoa,okspme,kaesprit}.
Other examples are keyboard clicking or pen dropping in
teleconference~\cite{georgiou1999alpha}, double-talk in echo
cancellation~\cite{vega2008new}, biological noise~\cite{Chitre2006}
or ice cracking~\cite{Bouvet1989compa} in various underwater
signals, out-of-band spectral leakage in CR~\cite{ZHU201594}, etc.
In such scenarios, the conventional algorithms like the dLMS
designed for Gaussian noise would undergo a significant performance
deterioration. To this end, many robust distributed algorithms have
been proposed. Some algorithms are based on the instantaneous
gradient-descent method to minimize different robust criteria, for
instance, the diffusion error nonlinearity
(dEN)~\cite{al2017robust}, diffusion least mean \emph{p}-th power
(dLMP) \cite{wen2013diffusion}, diffusion sign error LMS
(dSE-LMS)~\cite{ni2016diffusion}, and diffusion maximum
correntropy~\cite{ma2016diffusion} algorithms. Moreover, because of
the insensitivity of correntropy to impulsive noise, the maximum
total correntropy diffusion algorithm was proposed in~\cite{8405555}
for the case of large outliers in communication links. Nevertheless,
their main limitation is slow convergence especially when the nodes'
input signals are colored (highly correlated). As shown
in~\cite{al2017robust}, the dEN algorithm converges slower than the
dSE-LMS algorithm. In \cite{chouvardas2011adaptive}, by resorting to
the adaptive projected subgradient method, a robust diffusion
algorithm was developed which projects the output errors onto
halfspaces defined by Huber's error function at each node, thereby
speeding up the convergence. However, the setting of the parameters
controlling the algorithm's robustness requires prior knowledge of
the noise distribution which is often unavailable.

It is well-known that due to the exponentially weighted least squares (EWLS) criterion, the diffusion recursive LS (dRLS) algorithm provides fast convergence even for colored signals ~\cite{cattivelli2008diffusion,vahidpour2017analysis}. By means of the alternating direction method of multipliers to solve the EWLS problem, Mateos~\emph{et~al.} proposed another type of distributed RLS algorithm \cite{mateos2009distributed}. Following this algorithm, to reduce computation and communication costs, its variants were presented by censoring observations with small innovations~\cite{wang2018decentralized}. Likewise, these algorithms might experience convergence issues in impulsive noise environments, because impulsive noise samples are directly involved in the adaptation through output errors of nodes. For the single-agent case, many works have proposed RLS algorithms robust against impulsive noise, e.g.,~\cite{vega2009fast,bhotto2011robust}. However, distributed RLS-based techniques that are robust to impulsive noise have not been well investigated. The study in~\cite{ma2016robust} develops the diffusion recursive least~\emph{p}-th power (dRLP) algorithm, while its robustness relies on the value of~$p$ as the dLMP does.

Analogous to the RLS, distributed RLS requires high computational complexity. Apart from this, it may also suffer from numerical instability due to accumulation of round-off errors in finite-precision implementations \cite{zakharov2008low}. Aiming to address these problems, an efficient alternative method is the dichotomous coordinate-descent (DCD) that solves a system of normal equations associated with the RLS-type algorithms~\cite{zakharov2004multiplication,zakharov2008low,liu2009architecture,zakharov2013dcd}. In particular, the DCD method only involves shift and addition operations, thus the DCD-based RLS algorithms reduce the computational cost and improve the numerical stability in contrast with the original RLS counterparts, whilst preserving comparable estimation performance. For this reason, reference~\cite{arablouei2013reduced} also explored the use of the DCD in distributed networks, and developed the DCD-dRLS algorithm. It is worth mentioning that, however, the development of the DCD-based algorithms in impulsive noise environments has not been studied in single nor multi -agent scenarios.
\subsection{Contributions}
The focus of this paper is on developing robust distributed RLS algorithms for scenarios with impulsive noise. Specifically, our contributions are listed as follows:

1) A robust dRLS (R-dRLS) algorithm is developed by extending the framework of \cite{vega2009fast} to multi-agent scenarios with a diffusion distributed strategy. To ensure that the proposed R-dRLS algorithm has good convergence performance after an abrupt change in the set of parameters to be estimated, we also propose a diffusion-based non-stationary control (NC) method.

2) Theoretical insights into the  mean square steady-state and evolution behaviors of the R-dRLS algorithm in impulsive noise environment are presented.

3) We employ the DCD method for developing recursions used in the adaptation step of the R-dRLS algorithm, resulting in the DCD-R-dRLS algorithm with similar learning performance. Remarkably, the DCD-R-dRLS algorithm brings a reduction in computational complexity; especially for shift structured input regressors, it reduces the order of complexity from $\mathcal{O}(M^2)$ to $\mathcal{O}(M)$, where $M$ is the length of the estimated vector.

4) Simulation examples are presented to demonstrate the performance of the proposed algorithms in impusive noise scenarios described by either Bernoulli-Gaussian (BG) or $\alpha$-stable processes.

In comparison to the preliminary results~\cite{YYu2018} related to this work, the current version is further developed due to the main contributions 2) and 3). We slightly improve the NC method by a smoothing operation as shown in \eqref{015}. Moreover, the effectiveness of the proposed algorithms are also verified in an application to distributed spectrum estimation.

This paper is organized as follows. In Section II, the estimation problem is described. The R-dRLS algorithm is derived in Section III. Analyses of its mean square behavior are presented in Section IV. In Section~V, we review the DCD algorithm and propose the DCD-R-dRLS algorithm. In Section VI, extensive simulations are presented to verify the proposed algorithms. Finally, conclusions are given in Section~VII.

\emph{Notation}: Throughout the paper, all vectors are column vectors. We use the parenthesis on $i$ to denote matrices and vectors, and the subscript on $i$ to denote scalars. The superscript $(\cdot)^T$ denotes the transpose, $\lVert \cdot \rVert_2$ denotes the $l_2$-norm of a vector, and $E\{\cdot\}$ denotes the expectation of random variables. We use $\text{col}\{\cdot\cdot\cdot\}$ to denote an enlarged column vector structured by stacking its columns on top of each other, $\text{diag}\{\cdot\cdot\cdot\}$ to yield a diagonal matrix with its arguments, and $\text{Tr}\{\cdot\}$ to denote the trace of a matrix. $\bm I_M$ is the identity matrix of size $M\times M$, $\otimes$ is the Kronecker product, and $\bm 1$ is the column vector of length $M$ with all entries being one. For symmetric matrices $\bm X$ and $\bm Y$, the notation $\bm X \geq \bm Y$ stands for $\bm X-\bm Y \geq 0$, meaning that the matrix difference $\bm X-\bm Y$ is positive semi-definite.

\section{Problem Formulation}
\begin{figure}[tb]
    \begin{center}
    \begin{overpic}[scale=0.8,angle=0]{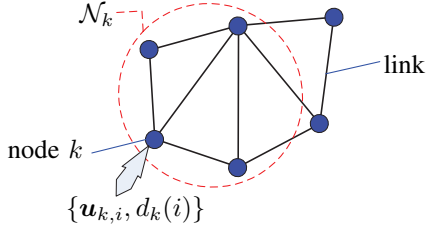}
        \put(6,55){\huge \normalsize $\mathcal{N}_k$}
        \put(-15,15){\huge \normalsize node $k$}
        \put(2,-2){\huge \normalsize $\left\lbrace \bm u_{k,i}, d_k(i)\right\rbrace $}
        \put(95,40){\huge \normalsize link}
    \end{overpic}
    \end{center}
    \hspace{2cm} \caption{\label{Fig1} A simple diffusion network showing a neighborhood $\mathcal{N}_k$ of node~$k$. At time instant $i$, node $k$ acquires the data $\{\bm u_{k,i},d_k(i)\}$.}
\end{figure}

Let us consider a diffusion network with $N$ nodes located at different positions in space, as shown in Fig. \ref{Fig1}, where each node communicates only with its neighboring nodes by a link (single-hop communication). All nodes connected directly to node $k$ (including itself) are referred to as its neighborhood, denoted as~$\mathcal{N}_k$. At every time instant $i\geq 0$, every node has access to an $M\times1$ input regressor vector $\bm u_{k,i}$ and an output measurement $d_k(i)$, which are related as:
\begin{equation}
d_k(i) = \bm u_{k,i}^T \bm w^o + v_k(i),
\label{001}
\end{equation}
where $\bm w^o$ is a parameter vector of size $M\times1$ to be estimated, and $v_k(i)$ is the additive noise at node \emph{k}. The additive noises $v_k(i)$ and $v_l(j)$ are spatially and temporally independent for $k\neq l$ and $i\neq j$. Moreover, any $\bm u_{k,i}$ is independent of any $v_l(j)$. The model (\ref{001}) is used in many applications~\cite{sayed2014adaptation,sayed2011adaptive}. The objective of the in-network processing is to estimate $\bm w^o$, using the available data $\{\bm u_{k,i},d_k(i)\}$ collected at nodes.
For this purpose, the global EWLS estimation problem is described as \cite{cattivelli2008diffusion}:
\begin{equation}
\label{002}
\begin{array}{rcl}
\begin{aligned}
&\bm w_i = \arg \min \limits_{\bm w} \\
&  \left\lbrace  \lambda^{i+1}\delta\lVert \bm w\rVert_2^2 + \sum\limits_{j=0}^i \lambda^{i-j} \sum\limits_{k=1}^N   \left(d_k(j)-\bm u_{k,i}^T\bm w\right)^2 \right\rbrace,
\end{aligned}
\end{array}
\end{equation}
where $\delta >0$ is a regularization constant, and $\lambda$ ($0< \lambda \leq1$) is the forgetting factor. The dRLS algorithm solves \eqref{002} in a diffusion-based distributed manner \cite{cattivelli2008diffusion}. As already mentioned in the introduction, the noise $v_k(i)$ may be non-Gaussian with impulsiveness so that the algorithms derived from~\eqref{002}, e.g., the dRLS algorithm, would exhibit poor convergence and even diverge. In general, when studying robust adaptive algorithms, both contaminated-Gaussian (CG) \cite{ni2016diffusion, bershad2008error} and $\alpha$-stable \cite{shao1993signal,CHEN2018318} random processes are often used for modeling impulsive noise.

\section{Proposed R-$\rm d$RLS Algorithm}
In this section, we derive the R-dRLS algorithm and propose a control method for endowing it with tracking capability. The diffusion strategy has two alternatives: the adapt-then-combine (ATC) and the combine-then-adapt (CTA). However, we focus only on the ATC policy, which performs first the adaptation step and then the combination step. This is based on the fact that the extension to CTA is straightforward by reversing the order of the adaptation and combination steps~\cite{sayed2014adaptive,sayed2013diffusion}. In what follows, we neglect the notation ATC for brevity.

\subsection{dRLS Algorithm}
To conveniently develop the R-dRLS algorithm, we re-derive here the dRLS algorithm from the following method instead of directly solving \eqref{002}.

In the adaptation step, every node $k$, at time instant $i$, finds an intermediate estimate $\bm \psi_{k,i}$ of $\bm w^o$ by minimizing the individual local cost function:
\begin{equation}
\label{003}
\begin{array}{rcl}
\begin{aligned}
J_k(\bm \psi_{k,i}) =& \lVert \bm \psi_{k,i}-\bm w_{k,i-1} \rVert_{\bm B_{k,i}}^2 \\
&+  [d_k(i)-\bm u_{k,i}^T\bm \psi_{k,i}]^2 ,
\end{aligned}
\end{array}
\end{equation}
with $\bm B_{k,i} =\bm \Phi_{k,i}-\bm u_{k,i}\bm u_{k,i}^T$, where
\begin{equation}
\label{004}
\begin{array}{rcl}
\begin{aligned}
\bm \Phi_{k,i} \triangleq &\lambda^{i+1}\delta \bm I_M + \sum\limits_{j=0}^i \lambda^{i-j}\bm u_{k,j}\bm u_{k,j}^T\\
=& \lambda \bm \Phi_{k,i-1} + \bm u_{k,i}\bm u_{k,i}^T
\end{aligned}
\end{array}
\end{equation}
is the time-averaged correlation matrix for the input vector at node $k$, and $\bm w_{k,i-1}$ is an estimate of $\bm w^o$ at node $k$ at time instant $i-1$. Notice that the quadratic form $\lVert \bm x \rVert_{\bm B_{k,i}}^2 \triangleq \bm x^T\bm B_{k,i}\bm x$ in \eqref{003} defines the Riemmanian distance between vectors $\bm \psi_{k,i}$ and $\bm w_{k,i-1}$, where $\bm B_{k,i}$ is a Riemannian metric tensor characterizing that the distance properties are not uniform along the $M$-dimensional space~\cite{amari1998natural,pelekanakis2014adaptive}.

Setting the derivative of $J_k(\bm \psi_{k,i})$ with respect to $\bm \psi_{k,i}$ to zero, we obtain
\begin{equation}
\label{005}
\begin{array}{rcl}
\begin{aligned}
\bm \psi_{k,i} = \bm w_{k,i-1} + \bm P_{k,i} \bm u_{k,i} e_k(i),
\end{aligned}
\end{array}
\end{equation}
where
\begin{equation}
\label{006}
\begin{array}{rcl}
\begin{aligned}
e_k(i) = d_k(i)-\bm u_{k,i}^T \bm w_{k,i-1}
\end{aligned}
\end{array}
\end{equation}
stands for the output error at node $k$, and
\begin{equation}
\label{007}
\begin{array}{rcl}
\begin{aligned}
\bm P_{k,i} &\triangleq \bm \Phi_{k,i}^{-1}\\
&= \frac{1}{\lambda} \left( \bm P_{k,i-1} - \frac{\bm P_{k,i-1}\bm u_{k,i} \bm u_{k,i}^T\bm P_{k,i-1}}{\lambda +\bm u_{k,i}^T \bm P_{k,i-1} \bm u_{k,i}} \right),
\end{aligned}
\end{array}
\end{equation}
with $\bm P_{k,i}$ initialized as $\left. \bm P_{k,0}=\delta^{-1} \bm I_M\right.$. The recursion~\eqref{007} is the result of applying the matrix inversion lemma~\cite{sayed2011adaptive}.

At the combination step, the intermediate estimates $\bm \psi_{m,i}$, $\left. m\in\mathcal{N}_k\right. $ from the neigborhood of node $k$ are linearly weighted, yielding a combined estimate $\bm w_{k,i}$ \cite{sayed2014adaptation}:
\begin{equation}
\label{008}
\begin{array}{rcl}
\begin{aligned}
\bm w_{k,i} = \sum\limits_{m\in\mathcal{N}_k}c_{m,k} \bm \psi_{m,i},
\end{aligned}
\end{array}
\end{equation}
where the combination coefficients $\{c_{m,k}\}$ are non-negative, and satisfy:
\begin{equation}
\label{009}
\sum\limits_{m\in\mathcal{N}_k}c_{m,k} =1 \text{, and } c_{m,k}=0 \text{ if } m\notin\mathcal{N}_k.
\end{equation}
Note that $c_{m,k}$ is a weight that node $k$ assigns to the intermediate estimate $\bm \psi_{m,i}$ received from its neighbor node $m$. If one assumes $\bm w_{k,i-1}=\bm \Phi_{k,i-1}^{-1} \bm z_{k,i-1}$ in \eqref{005}, where $\bm z_{k,i} = \lambda \bm z_{k,i-1} + \bm u_{k,i} d_k(i)$, \eqref{005} is a standard RLS update for node~$k$. In summary, \eqref{005}-\eqref{008} formulate the dRLS algorithm. It is noteworthy that the term $\bm P_{k,i} \bm u_{k,i}$ in \eqref{005} provides the decorrelating ability for colored inputs, thus speeding up the convergence.

\textbf{Remark 1}: In general, $\{c_{m,k}\}$ in \eqref{009} are determined by one of many static rules (e.g., the Metropolis rule \cite{takahashi2010diffusion} that we adopt in this paper) which keeps them constant during the estimation. Considering that nodes may be working under different signal-to-noise ratios (SNRs), several adaptive rules have been proposed to optimize the algorithm behavior \cite{takahashi2010diffusion,6854838,7936509}. However, these adaptive rules are severely polluted when impulsive noise samples appear, since the output errors at nodes directly participate in the adaptation of $c_{m,k}$. Designing a robust adaptive rule is an alternative, but it is not the focus of this paper. In another approach, based on the detection of impulsive noise, Ahn \emph{et~al.} proposed a robust variable weighting coefficients dLMS (RVWC-dLMS) algorithm which sets the weighting coefficients to zero at nodes disturbed by impulsive noise \cite{ahn2017new}. Likewise, the RVWC scheme can be extended to dRLS in a straightforward way, resulting in the RVWC-dRLS algorithm with robustness in impulsive noises\footnote{In the literature, the RVWC scheme was presented for more general diffusion strategies (namely, also exchanging information among nodes in adaptation step). However, here we do not consider this case for a fair comparison. Besides, such general diffusion strategies require higher computational complexity and communication load \cite{sayed2013diffusion}.}, as can be seen in the simulations later on.

\subsection{R-dRLS Algorithm}
An impulsive noise sample at time instant $i$ might lead the dRLS algorithm to diverge via $e_k(i)$ in \eqref{005} due to its large amplitude and the propagation of its effect. This degradation effect can last for many iterations. To endow the algorithm with robustness in impulsive noise scenarios, we propose to minimize \eqref{003} under the following constraint:
\begin{equation}
\label{010}
\begin{array}{rcl}
\begin{aligned}
\lVert \bm \psi_{k,i}-\bm w_{k,i-1}\rVert_2^2 \leq \xi_k(i-1),
\end{aligned}
\end{array}
\end{equation}
where $\xi_k(i-1)$ is a positive bound. A similar constraint appeared in an adaptive filter for a single agent scenario \cite{vega2009fast}, but when generalizing to the distributed version with multiple agents, the constraint could be imposed on the adaptation at all the nodes. This constraint represents that the energy (squared norm) of the update at every node $k$ from $\bm w_{k,i-1}$ to $\bm \psi_{k,i}$ always does not exceed the amount $\xi_k(i-1)$ regardless of the type of noise (possibly, impulsive noise), thereby guaranteeing the robustness of the algorithm. In doing so, if~\eqref{005} satisfies~\eqref{010}, i.e.,
\begin{equation}
\label{011}
\begin{array}{rcl}
\begin{aligned}
\lVert \bm g_{k,i} \rVert_2 \lvert e_k(i)\rvert \leq \sqrt{\xi_k(i-1)},
\end{aligned}
\end{array}
\end{equation}
where $\left. \bm g_{k,i}\triangleq \bm P_{k,i} \bm u_{k,i}\right. $ represents the Kalman gain vector, then \eqref{005} is a solution of the above constrained minimization problem. Conversely, if \eqref{010} is not satisfied (usually in the case of appearance of impulsive noise), i.e., $\left. \lVert \bm g_{k,i} \rVert_2 \lvert e_k(i)\rvert > \sqrt{\xi_k(i-1)}\right. $, we propose to replace the update \eqref{005} by its normalized form to satisfy the equality in \eqref{010}, which is described by
\begin{equation}
\label{012}
\begin{array}{rcl}
\begin{aligned}
\bm \psi_{k,i} = \bm w_{k,i-1} + \sqrt{\xi_k(i-1)} \frac{\bm g_{k,i}}{\lVert\bm g_{k,i}\rVert_2}\text{sign}(e_k(i)),\\
\end{aligned}
\end{array}
\end{equation}
where $\text{sign}(\cdot)$ is the sign function. Thus, combining \eqref{005}, \eqref{011} and \eqref{012}, we obtain the adaptation step for each node $k$ as:
\begin{equation}
\label{013}
\begin{array}{rcl}
\begin{aligned}
\bm \psi_{k,i} = \bm w_{k,i-1} + \min \left[  \frac{\sqrt{\xi_k(i-1)}}{\lVert \bm g_{k,i} \rVert_2 \lvert e_k(i)\rvert},\; 1 \right] \bm g_{k,i} e_k(i).\\
\end{aligned}
\end{array}
\end{equation}

Evidently, the crucial problem is how to properly choose the bound $\xi_k(i)$ as it controls the robustness of the algorithm against impulsive noise and influences its dynamic behavior.  To be more specific, we wish $\xi_k(i)$ to have larger values at the earlier adaptation stage to provide a fast initial convergence, while for enforcing good robustness against impulsive noise, its values cannot be too large. In addition, we also wish to obtain a small estimation error at steady-state, so $\xi_k(i)$ should be reduced to a small value. Based on these requirements, we consider the equality in \eqref{010} to propose a useful recursive method for adjusting $\xi_k(i)$, as described by
\begin{equation}
\label{014}
\begin{array}{rcl}
\begin{aligned}
\zeta_k(i) =& \beta \xi_k(i-1) + (1-\beta) \left\| \bm \psi_{k,i}-\bm w_{k,i-1}\right\|_2^2 \\
=\beta  &\xi_k(i-1) + (1-\beta) \min [\lVert \bm g_{k,i}\rVert_2^2 e_k^2(i),\xi_k(i-1) ], \\
\xi_k(i) =&\sum \limits_{m\in\mathcal{N}_k}c_{m,k} \zeta_m(i),
\end{aligned}
\end{array}
\end{equation}
where $\beta$ is a memory factor with $0<\beta<1$. At every node~$k$, $\xi_k(i)$ can be initialized by $\xi_k(0)= E_c\sigma_{d,k}^2/(M\sigma_{u,k}^2)$, where $E_c$ is a positive integer, and $\sigma_{d,k}^2$ and $\sigma_{u,k}^2$ are powers of the output measurement $d_k(i)$ and the input regressor $\bm u_{k,i}$, respectively.
As one can see in~\eqref{014}, every node $k$ not only uses its own adaptive rule to update $\xi_k(i)$, but also exploits the side information $\zeta_m(i)$ transmitted from its neighboring nodes by the diffusion cooperation. In doing so, the proposed R-dRLS algorithm is more effective at computing consistent estimates at all nodes, which will be observed in Section VI-A. Table~\ref{table_1} details the proposed R-dRLS algorithm together with the NC method.
\begin{table}[tbp]
    \scriptsize
    \centering
    \caption{ Proposed R-$\rm d$RLS Algorithm Allied with the NC Method}
    \label{table_1}
    \begin{tabular}{lc}
        \hline
        \text{Parameters:} $0< \beta <1$, $\lambda$, $\delta$ and $E_c$ (R-dRLS part); \\
        \;\;\;\;\;\;\;\;\;\;\;\;\;\;\;\;\;\;$\varrho$, $\tau$ and $t_\text{th}$ (NC part)\\
        \hline
        \text{Initialization}: $\bm w_{k,0} = \bm 0$, $\bm P_{k,0}=\delta^{-1} \bm I_M$ and $\xi_k(0)= E_c \frac{\sigma_{d,k}^2}{M\sigma_{u,k}^2}$ (R-dRLS part);\\ \;\;\;\;\;\;\;\;\;\;\;\;\;\;\;\;\;\;\;\;$\varTheta_{\text{old},k}=\varTheta_{\text{new},k}=0$, $\sigma_{e,k}^2=0$, $V_t=\varrho M$\\
        \;\;\;\;\;\;\;\;\;\;\;\;\;\;\;\;\;\;\;\; and $V_d = 0.75V_t$ (NC part)\\
        \hline
        \textbf{for} iteration $i=1, 2, 3,...$ \\
        \;\;\textbf{for} node  $k=1, 2, 3,...,N$ \\
        \;\;\;\;[R-dRLS part:]\\
        \;\;\;\;$e_k(i) = d_k(i)-\bm u_{k,i}^T \bm w_{k,i-1}$ \\
        \;\;\;\;$\bm P_{k,i} = \frac{1}{\lambda} \left( \bm P_{k,i-1} - \frac{\bm P_{k,i-1}\bm u_{k,i} \bm u_{k,i}^T\bm P_{k,i-1}}{\lambda +\bm u_{k,i}^T \bm P_{k,i-1} \bm u_{k,i}} \right) $ \\
        \;\;\;\;$\bm g_{k,i} = \bm P_{k,i}\bm u_{k,i}$\\
        \;\;\;\;$\bm \psi_{k,i} = \bm w_{k,i-1} + \min \left[  \frac{\sqrt{\xi_k(i-1)}}{\lVert \bm g_{k,i} \rVert_2 \lvert e_k(i)\rvert},\; 1 \right]  \bm g_{k,i} e_k(i)$\\
        \;\;\;\;$\bm w_{k,i} = \sum\limits_{m\in\mathcal{N}_k}c_{m,k} \bm \psi_{m,i}$\\
        \;\;\;\;[NC part:] \\
        \;\;\;\;\emph{Step 1: to compute} $\varDelta_k(i)$\\
        \;\;\;\;\textbf{if}\;$i=nV_t, n=0,1,2,...$\\
        \;\;\;\;\;\;$\bm a_{k,i}^T = \mathcal{R} \left( \left[ \frac{e_k^2(i)}{\|\bm u_{k,i}\|_2^2},\frac{e_k^2(i-1)}{\|\bm u_{k,i-1}\|_2^2},\text{...}, \frac{e_k^2(i-V_t+1)}{\|\bm u_{k,i-V_t+1}\|_2^2} \right] \right) $\\
        \;\;\;\;\;\;$\sigma_{e,k}^2 \leftarrow \tau \sigma_{e,k}^2 + (1-\tau)\bm a_{k,i}^T \bm e$\\
        \;\;\;\;\;\;$\varTheta_{\text{new},k} = \frac{1}{V_t-V_d}\sum\limits_{m\in\mathcal{N}_k}c_{m,k} \sigma_{e,m}^2$\\
        \;\;\;\;\;\;$\varDelta_k(i) = \frac{\varTheta_{\text{new},k}-\varTheta_{\text{old},k}}{\xi_k(i-1)}$\\
        \;\;\;\;\textbf{end}\\
        \;\;\;\;\emph{Step 2: to reset} $\xi_k(i)$\\
        \;\;\;\;\textbf{if} $\varDelta_k(i)> t_\text{th} $\\
        \;\;\;\;$\zeta_k(i) = \xi_k(0)$, $\bm P_{k,i}= \bm P_{k,0}$\\
        \;\;\;\;\textbf{elseif} \; $\varTheta_{\text{new},k}>\varTheta_{\text{old},k}$\\
        \;\;\;\;\;\;$\zeta_k(i) = \xi_k(i-1) + (\varTheta_{\text{new},k}-\varTheta_{\text{old},k})$\\
        \;\;\;\;\textbf{else}\\
        \;\;\;\;\;\;$\zeta_k(i) = \beta \xi_k(i-1) + (1-\beta) \min \left[ \lVert \bm g_{k,i}\rVert_2^2 e_k^2(i),\;\xi_k(i-1) \right]$\\
        \;\;\;\;\textbf{end}\\
        \;\;\;\;$\varTheta_{\text{old},k} = \varTheta_{\text{new},k}$\\
        \;\;\;\;$\xi_k(i) =\sum\limits_{m\in\mathcal{N}_k}c_{m,k} \zeta_m(i)$\\
        \;\;\textbf{end}\\
        \textbf{end}\\
        \hline
    \end{tabular}
\end{table}

\textbf{Remark 2}: As can be seen from \eqref{013}, the operation mode of the proposed R-dRLS algorithm in the adaptation step can be as follows. At time instant $i$, if $\lVert \bm g_{k,i}\rVert_2^2 e_k^2(i) \leq \xi_k(i-1)$, the classical RLS update is performed. If not, the squared norm of the RLS increment is limited to the amount $\xi_k(i-1)$ as in~\eqref{012} for guaranteeing the robustness in impulsive noise. At the early iterations, the values of $\xi_k(i-1)$ can be high compared to $\lVert \bm g_{k,i}\rVert_2^2e_k^2(i)$ so that the algorithm will behave as the dRLS algorithm, providing a fast initial convergence. On the other hand, whenever an impulsive noise sample appears, due to its significant magnitude, the R-dRLS algorithm will work as an dRLS update multiplied by a very small scaling factor $\frac{\sqrt{\xi_k(i-1)}}{\lVert \bm g_{k,i}\rVert_2 |e_k(i)|}$. It has been shown in~\cite{song2013normalized, hur2017variable} that in the adaptation update term, the multiplication of a small scaling factor can reduce the negative influence of impulsive noise on the estimation. Thus, this also indirectly implies that the R-dRLS algorithm has robustness against impulsive noise. Moreover, the robustness is further maintained over the iterations, due to the decreasing property of $\xi_k(i)$ given by~\eqref{014}. In addition to this, the diminishing $\xi_k(i)$ also leads to a reduction in the steady-state error of the algorithm. To sum up, the R-dRLS algorithm can be considered as an improved dRLS algorithm with a variable 'step-size' scheme which has an automatic switch between 1 and $\frac{\sqrt{\xi_k(i-1)}}{\lVert \bm g_{k,i}\rVert_2 |e_k(i)|}$, as can be observed in~\eqref{013}.

\subsection{NC Method}
As a consequence of the diminishing sequence $\{\xi_k(i)\}$, the R-dRLS algorithm has poor ability of tracking (i.e., re-convergence of the algorithm) after $\bm w^o$ undergoes an abrupt change. In order to overcome this problem, inspired by the idea in~\cite{vega2008new} for the single-agent scenario, we propose here a diffusion-based NC method, as summarized in Table~\ref{table_1}. The NC method is implemented in the following two steps.

\emph{Step 1}: A variable $\varDelta_k(i)$ at node $k$ is computed once for every $V_t$ iterations, to judge whether the unknown vector changed or not. In this step, $\bm a_{k,i}^T = \mathcal{R} \left( \left[ \frac{e_k^2(i)}{\|\bm u_{k,i}\|_2^2},\frac{e_k^2(i-1)}{\|\bm u_{k,i-1}\|_2^2},\text{...}, \frac{e_k^2(i-V_t+1)}{\|\bm u_{k,i-V_t+1}\|_2^2} \right] \right)$ with $\mathcal{R}(\cdot)$ denoting the ascending arrangement for its arguments. With $\bm e=[1,...,1,0,...,0]^T$ being a vector whose first $V_t-V_d$ elements set to one, where $V_d$ is a positive integer with $V_d<V_t$, the product $\bm a_{k,i}^T \bm e$ can remove the effect of outliers (e.g., impulsive noise samples) when computing $\varDelta_k(i)$. We use a smooth filtering of $\bm a_{k,i}^T \bm e$ to avoid large fluctuations in computing $\varTheta_{\text{new},k}$ (see Table \ref{table_1}), as follows:
\begin{equation}
\label{015}
\sigma_{e,k}^2 \leftarrow \tau \sigma_{e,k}^2 + (1-\tau)\bm a_{k,i}^T \bm e,
\end{equation}
where $\tau$, $0< \tau < 1$, is a memory factor. Note that, every node $k$ to compute $\varTheta_{\text{new},k}$ also combines the information from its neighboring nodes based on a diffused cooperation; $\varTheta_{\text{old},k}$ stores the value of $\varTheta_{\text{new},k}$ at the last time instant.

From \emph{Step 1}, one can see that using a larger $V_t$, the algorithm has lower steady-state error but a higher delay in tracking. Moreover, for a large occurrence probability of impulsive noise, the value of~$V_d$ should be increased to better discard the impulsive noise samples in the computation of $\varDelta_k(i)$. From our extensive simulations, we found out that for both $V_t$ and $V_d$, good choices are $V_t=\varrho M$ with $1\leq \varrho \leq 3$ and $V_d=0.75V_t$~\cite{vega2008new}.

\emph{Step 2}: If $\varDelta_k(i) > t_\text{th}$, where $t_\text{th}$ is a predefined threshold, it is decided that a change of $\bm w^o$ has occurred. Then, we reset $\xi_k(i)$ to its initial value $\xi_k(0)$ so that the R-dRLS algorithm can track this change rapidly. Meanwhile, $\bm P_{k,i}$ should also be re-initialized with $\bm P_{k,0}$.

It is worth noting that in this scheme the parameters $\tau,\;\varrho$, and $t_\text{th}$ are not affected by each other so that their choices are simplified.

\section{Mean Square Performance Analyses}
\subsection{Steady-state Behavior}
In this section, we discuss the steady-state behavior of the R-dRLS algorithm in impulsive noise. Assuming that the vector $\bm w^o$ is invariant, then we define the estimate deviation and intermediate estimate deviation vectors respectively as:
\begin{equation}
\label{016}
\begin{array}{rcl}
\begin{aligned}
\widetilde{\bm w}_{k,i} &\triangleq \bm w^o - \bm w_{k,i},\\
\widetilde{\bm \psi}_{k,i} &\triangleq \bm w^o - \bm \psi_{k,i}.
\end{aligned}
\end{array}
\end{equation}
Using these definitions and \eqref{014}, it is easy to rearrange \eqref{013} and \eqref{008}, respectively, as:
\begin{equation}
\label{017}
\begin{array}{rcl}
\begin{aligned}
\widetilde{\bm \psi}_{k,i} = \widetilde{\bm w}_{k,i-1} -\sqrt{\frac{\zeta_k(i)-\beta\xi_k(i-1)}{1-\beta}} \frac{\bm g_{k,i}}{\lVert \bm g_{k,i} \rVert_2} \text{sign}(e_k(i)),
\end{aligned}
\end{array}
\end{equation}
and
\begin{equation}
\label{018}
\begin{array}{rcl}
\begin{aligned}
\widetilde{\bm w}_{k,i} = \sum\limits_{m\in\mathcal{N}_k}c_{m,k} \widetilde{\bm \psi}_{m,i}.
\end{aligned}
\end{array}
\end{equation}

Equating the squared $l_2$-norm of both sides of \eqref{017} and then taking the expectation, we obtain
\begin{equation}
\label{019}
\begin{array}{rcl}
\begin{aligned}
E&\left\lbrace \lVert \widetilde{\bm \psi}_{k,i} \rVert_2^2 \right\rbrace = E\left\lbrace \lVert \widetilde{\bm w}_{k,i-1} \rVert_2^2 \right\rbrace\\
&-2E\left\lbrace \sqrt{\frac{\zeta_k(i)-\beta\xi_k(i-1)}{1-\beta}} \frac{\widetilde{\bm w}_{k,i-1}^T \bm g_{k,i}}{\lVert \bm g_{k,i} \rVert_2} \text{sign}(e_k(i))\right\rbrace\\
&+\frac{E\left\lbrace \zeta_k(i)\right\rbrace -\beta E\left\lbrace \xi_k(i-1)\right\rbrace }{1-\beta}.
\end{aligned}
\end{array}
\end{equation}
Likewise treating \eqref{018} and applying Jensen's inequality~\cite[p.77]{boyd2004convex}, we obtain
\begin{equation}
\label{020}
\begin{array}{rcl}
\begin{aligned}
E\left\lbrace \lVert \widetilde{\bm w}_{k,i}\rVert_2^2 \right\rbrace \leq \sum\limits_{m\in\mathcal{N}_k}c_{m,k} E\left\lbrace \lVert \widetilde{\bm \psi}_{m,i}\rVert_2^2\right\rbrace.
\end{aligned}
\end{array}
\end{equation}

Typically, $\beta$ is close to 1 so that the variances of $\zeta_k(i)$ and $\xi_k(i)$ given in \eqref{014} would be small enough. Accordingly, it can be assumed that
\begin{equation}
\label{021}
\begin{array}{rcl}
\begin{aligned}
&E\left\lbrace \sqrt{\frac{\zeta_k(i)-\beta\xi_k(i-1)}{1-\beta}} \frac{\widetilde{\bm w}_{k,i-1}^T \bm g_{k,i}}{\lVert \bm g_{k,i} \rVert_2} \text{sign}(e_k(i))\right\rbrace \approx\\
&\sqrt{\frac{E\left\lbrace \zeta_k(i)\right\rbrace -\beta E\left\lbrace \xi_k(i-1)\right\rbrace }{1-\beta}} E\left\lbrace \frac{\widetilde{\bm w}_{k,i-1}^T \bm g_{k,i}}{\lVert \bm g_{k,i} \rVert_2} \text{sign}(e_k(i))\right\rbrace.
\end{aligned}
\end{array}
\end{equation}
Then, with this approximation, \eqref{019} is changed to
\begin{equation}
\label{022}
\begin{array}{rcl}
\begin{aligned}
E\left\lbrace \lVert \widetilde{\bm \psi}_{k,i} \rVert_2^2 \right\rbrace = & E\left\lbrace \lVert \widetilde{\bm w}_{k,i-1} \rVert_2^2 \right\rbrace-\\
&2\sqrt{\frac{E\left\lbrace \zeta_k(i)\right\rbrace - \beta E\left\lbrace \xi_k(i-1)\right\rbrace }{1-\beta}} \times \\
&\underbrace{E\left\lbrace \frac{\widetilde{\bm w}_{k,i-1}^T \bm g_{k,i}}{\lVert \bm g_{k,i} \rVert_2} \text{sign}(e_k(i))\right\rbrace} \limits_{(a)} +\\
&\frac{E\left\lbrace \zeta_k(i)\right\rbrace - \beta E\left\lbrace \xi_k(i-1)\right\rbrace }{1-\beta}.
\end{aligned}
\end{array}
\end{equation}

To deal with the (a) term in \eqref{022}, some assumptions are helpful.

\textbf{Assumption 1:} The input regressors $\bm u_{k,i}$ are zero-mean with covariance matrices $\bm R_k=E\{\bm u_{k,i}\bm u_{k,i}^T\}$ and spatially independent.

\textbf{Assumption 2:} The regressors $\{\bm u_{k,i}\}$ are independent of the estimates $\{\bm w_{m,j}\}$ for $j\leq i$ and all $k,m$, referred to as the \emph{independence assumption}, which is known as useful in the analysis of adaptive algorithms \cite{sayed2011adaptive} and distributed estimation algorithms \cite{sayed2014adaptation,chen2013distributed}.

\textbf{Assumption 3:} There is an iteration number $i_0$ such that for all $i>i_0$, the time-averaged matrix $\bm \Phi_{k,i}$ at every node $k$ can be replaced by its expected value $E\left\lbrace \bm \Phi_{k,i} \right\rbrace$. This is an ergodicity assumption since $0\ll \lambda<1$, and from \eqref{004} we have
\begin{equation}
\label{023}
\begin{array}{rcl}
\begin{aligned}
\lim_{i\rightarrow \infty} E\left\lbrace \bm \Phi_{k,i} \right\rbrace = \frac{\bm R_{k}}{1-\lambda} \triangleq \bar{\bm \Phi}_k.
\end{aligned}
\end{array}
\end{equation}
Correspondingly, we can also replace the random matrix $\bm \Phi_{k,i}^{-1}$ by $\bar{\bm \Phi}_k^{-1}\triangleq E\left\lbrace \bm \Phi_{k,i}^{-1}\right\rbrace $ for a sufficiently large number of iterations $i$. Note that such replacements are commonly used in the performance analysis of RLS-type algorithms, see~\cite{cattivelli2008diffusion,sayed2011adaptive,zhang2017variable,vahidpour2017analysis} and the references therein.

Applying assumption 3, we are able to represent the term (a) in \eqref{022} as:
\begin{equation}
\label{024}
\begin{array}{rcl}
\begin{aligned}
&E\left\lbrace \frac{\widetilde{\bm w}_{k,i-1}^T \bm g_{k,i}}{\lVert \bm g_{k,i} \rVert_2} \text{sign}(e_k(i))\right\rbrace \\
&\;\;\;\;\;\;\;\;\;\;\approx E\left\lbrace \frac{\widetilde{\bm w}_{k,i-1}^T \bar{\bm \Phi}_k^{-1} \bm u_{k,i}}{\sqrt{\bm u_{k,i}^T \bar{\bm \Phi}_k^{-2} \bm u_{k,i}}} \text{sign}(e_k(i))\right\rbrace\\
&\;\;\;\;\;\;\;\;\;\; = E\left\lbrace \frac{\widetilde{\bm w}_{k,i-1}^T \bm R_k^{-1} \bm u_{k,i}}{\sqrt{\bm u_{k,i}^T \bm R_k^{-2} \bm u_{k,i}}} \text{sign}(e_k(i))\right\rbrace.
\end{aligned}
\end{array}
\end{equation}
In the light of assumption~1, if the dimension of $\bm w^o$ is large, i.e., $M\gg1$,  the fluctuation of the denominator term in~\eqref{024} from one iteration to the next can be assumed to be small. So, we could make the following approximation (which is also verified in Appendix A):
\begin{equation}
\label{025}
\begin{array}{rcl}
\begin{aligned}
&E\left\lbrace \frac{\widetilde{\bm w}_{k,i-1}^T \bm R_k^{-1} \bm u_{k,i}}{\sqrt{\bm u_{k,i}^T \bm R_k^{-2} \bm u_{k,i}}} \text{sign}(e_k(i))\right\rbrace \approx\\
&\;\;\;\;\;\;\;\;\;\; \chi_k E\left\lbrace e_{a,k}^{\bm R_k^{-1}}(i) \text{sign}\left( e_{a,k}(i)+v_k(i) \right) \right\rbrace,
\end{aligned}
\end{array}
\end{equation}
where
\begin{equation}
\label{026}
\begin{array}{rcl}
\begin{aligned}
e_{a,k}(i) &\triangleq \widetilde{\bm w}_{k,i-1}^T \bm u_{k,i}, \\
e_{a,k}^{\bm R_k^{-1}}(i) &\triangleq \widetilde{\bm w}_{k,i-1}^T \bm R_k^{-1} \bm u_{k,i}, \\
\chi_k &= E\left\lbrace \frac{1}{\sqrt{\bm u_{k,i}^T \bm R_k^{-2} \bm u_{k,i}}} \right\rbrace.
\end{aligned}
\end{array}
\end{equation}

Considering the presence of impulsive noise, we need the following assumptions to continue the analysis.

\textbf{Assumption 4:} At every node $k$, the additive noise $v_k(i)$ is drawn from a CG random process, $v_k(i)=\theta_k(i)+\eta_k(i)$, where $\theta_k(i)$ is the background noise assumed to be zero-mean white Gaussian with variance $\sigma_{\theta,k}^2$. The impulsive part $\eta_k(i)$ is described as $\eta_k(i)=b_k(i) g_k(i)$, where $b_k(i)$ is drawn from a Bernoulli random process with the probability $\left. P[b_k(i)=1]=p_{r,k}\right. $, and $g_k(i)$ is drawn from a white Gaussian random process with zero-mean and variance $\sigma_{g,k}^2=\hbar \sigma_{\theta,k}^2$, $\hbar\gg 1$. Usually, $p_{r,k}$ is also called the appearance probability of an impulsive noise sample.

Then, the mean and variance of $v_k(i)$ are zero and $\sigma_{v,k}^2 = p_{r,k}(\hbar+1)\sigma_{\theta,k}^2 + (1-p_{r,k})\sigma_{\theta,k}^2$, respectively. Note that, only when $p_{r,k}=0$ or 1, $v_k(i)$ is Gaussian; otherwise, $v_k(i)$ is non-Gaussian. Also, $v_k(i)$ conditioned on $b_k(i)$ is Gaussian~\cite{bershad2008error}. Although the $\alpha$-stable process is more appropriate for modeling impulsive noise in practice \cite{georgiou1999alpha, shao1993signal,CHEN2018318}, one would not consider it in the algorithms' analysis because its probability density function has no explicit form. Accordingly, the above assumption was used frequently for performance analysis of adaptive algorithms in impulsive noise environments, providing mathematical tractability~\cite{ni2016diffusion, vega2008new, zhou2011new,bershad2008error}.

Furthermore, as pointed out in \cite{al2003transient}, when $M\gg1$, then by using the central limit theorem, it can be assumed that $e_{a,k}(i)$ and $e_{a,k}^{\bm \Sigma}(i)$ are zero mean Gaussian variables for any constant matrix $\bm \Sigma$. Then, we can employ the following Lemma.

\emph{Lemma:}  Let $e_a$ and $u$ be jointly Gaussian zero-mean random variables. Let $e=e_a + v$, where $v$ is a zero-mean CG random variable with variance $\sigma_{v}^2 = p_{r}(\hbar+1)\sigma_{\theta}^2 + (1-p_{r})\sigma_{\theta}^2$, and $v$ is independent of $e_a$ and $u$. If $e_1=e_a + \omega_1$ and $e_2=e_a + \omega_2$, where $\omega_1$ and $\omega_2$ are zero-mean Gaussian random variables with variances $\sigma_{\omega_1}^2=(\hbar+1)\sigma_{\theta}^2$ and $\sigma_{\omega_2}^2= \sigma_{\theta}^2$, and are independent of $u$ and $e_a$, then
\begin{equation*}
\begin{array}{rcl}
\begin{aligned}
E\{\text{sign}(e)u\} = p_{r}E\{\text{sign}(e_1)u\} +(1-p_{r}) E\{\text{sign}(e_2)u\}.
\end{aligned}
\end{array}
\end{equation*}
Such a Lemma has been commonly used in the past for analyzing the sign-based algorithms \cite{zhou2011new,ni2016diffusion}.
Based on Price's theorem in \cite{price1958useful}, Lemma and assumption 2, we can establish the following equation
\begin{equation}
\label{027}
\begin{array}{rcl}
\begin{aligned}
E\left\lbrace \left. e_{a,k}^{\bm R_k^{-1}}(i) \text{sign}\left( e_{a,k}(i)+v_k(i) \right) \right|  \widetilde{\bm w}_{k,i-1} \right\rbrace\\
= \varpi_{k,i} E\left\lbrace \left. e_{a,k}^{\bm R_k^{-1}}(i) e_{a,k}(i) \right|   \widetilde{\bm w}_{k,i-1} \right\rbrace,
\end{aligned}
\end{array}
\end{equation}
where
\begin{equation}
\begin{array}{rcl}
\begin{aligned}
\label{028}
\varpi_{k,i} = \sqrt{\frac{2}{\pi}} \left\lbrace \frac{p_{r,k}}{\sqrt{E\{e_{a,k}^2(i)\}+(\hbar+1)\sigma_{\theta}^2\}}} \right.&  \\
+\left. \frac{1-p_{r,k}}{\sqrt{E\{e_{a,k}^2(i)\}+\sigma_{\theta}^2\}}} \right\rbrace \neq 0 & ,
\end{aligned}
\end{array}
\end{equation}
and the notation $E\{s|q\}$ accounts for the expectation of~$s$ conditioned on $q$.
Subsequently, the right-hand term in equality~\eqref{025} becomes
\begin{equation}
\label{029}
\begin{array}{rcl}
\begin{aligned}
E&\left\lbrace e_{a,k}^{\bm R_k^{-1}}(i) \text{sign}\left( e_{a,k}(i)+v_k(i) \right) \right\rbrace\\
&=E\left\lbrace E\left\lbrace \left. e_{a,k}^{\bm R_k^{-1}}(i) \text{sign}\left( e_{a,k}(i)+v_k(i) \right) \right|  \widetilde{\bm w}_{k,i-1} \right\rbrace \right\rbrace \\
&= \varpi_{k,i} E\left\lbrace e_{a,k}^{\bm R_k^{-1}}(i) e_{a,k}(i)  \right\rbrace\\
&\stackrel{(a)}{=} \varpi_{k,i} E\left\lbrace \lVert \widetilde{\bm w}_{k,i-1} \rVert_2^2\right\rbrace ,
\end{aligned}
\end{array}
\end{equation}
where the equality (a) is the result of using \eqref{026} under assumption 2.

Substituting \eqref{025} and \eqref{029} into \eqref{022}, it is rearranged as
\begin{equation}
\label{030}
\begin{array}{rcl}
\begin{aligned}
&E\left\lbrace \lVert \widetilde{\bm \psi}_{k,i} \rVert_2^2 \right\rbrace = E\left\lbrace \lVert \widetilde{\bm w}_{k,i-1} \rVert_2^2 \right\rbrace\\
&-2\chi_k \varpi_{k,i} \sqrt{\frac{E\left\lbrace \zeta_k(i)\right\rbrace - \beta E\left\lbrace \xi_k(i-1)\right\rbrace }{1-\beta}} E\left\lbrace \lVert \widetilde{\bm w}_{k,i-1} \rVert_2^2\right\rbrace \\
&+\frac{E\left\lbrace \zeta_k(i)\right\rbrace - \beta E\left\lbrace \xi_k(i-1)\right\rbrace }{1-\beta}.
\end{aligned}
\end{array}
\end{equation}

Next, we introduce the following network global vectors:
\begin{equation}
\begin{array}{rcl}
\begin{aligned}
\label{031}
 {\mathcal{\bm X}}_i&\triangleq\text{col} \left\lbrace  E\{\| \widetilde{\bm \psi}_{1,i} \|_2^2\} , ..., E\{ \| \widetilde{\bm \psi}_{N,i} \|_2^2\}\right\rbrace , \\
{\mathcal{\bm W}}_i&\triangleq\text{col} \left\lbrace  E\{\| \widetilde{\bm w}_{1,i} \|_2^2\} , ..., E\{\| \widetilde{\bm w}_{N,i} \|_2^2\}\right\rbrace , \\
\end{aligned}
\end{array}
\end{equation}
and the network global matrices
\begin{equation}
\begin{array}{rcl}
\begin{aligned}
\label{032}
\bm \Lambda_i &\triangleq \text{diag} \left\lbrace \chi_1\varpi_{1,i}, ..., \chi_N\varpi_{N,i} \right\rbrace, \\
\bm \Omega_i &\triangleq \text{diag}\left\lbrace \sqrt{\frac{E\left\lbrace \zeta_1(i)\right\rbrace - \beta E\left\lbrace \xi_1(i-1)\right\rbrace }{1-\beta}},...,\right. \\
&\;\;\;\;\;\;\;\;\;\;\;\;\;\;\;\left. \sqrt{\frac{E\left\lbrace \zeta_N(i)\right\rbrace - \beta E\left\lbrace \xi_N(i-1)\right\rbrace }{1-\beta}} \right\rbrace. \\
\end{aligned}
\end{array}
\end{equation}
Also, we define the matrix $\bm C$ to collect the combination coefficients, i.e., $[\bm C]_{m,k}=c_{m,k}$. Following \eqref{031} and \eqref{032}, we can formulate \eqref{020} and \eqref{030} for all nodes as follows:
\begin{equation}
\begin{array}{rcl}
\begin{aligned}
\label{033}
\mathcal{\bm W}_i &\leq \bm C^T {\mathcal{\bm X}}_i \\
&= \bm C^T \left[ \mathcal{\bm W}_{i-1} - 2\bm \Lambda_i \bm \Omega_i \mathcal{\bm W}_{i-1} + \bm \Omega_i^2 \bm 1_M \right].\\
\end{aligned}
\end{array}
\end{equation}
Taking the $\infty$-norm for both sides of \eqref{033} leads to
\begin{equation}
\begin{array}{rcl}
\begin{aligned}
\label{034}
\lVert \mathcal{\bm W}_i \rVert_\infty &\leq \left\|  \bm C^T \left( \mathcal{\bm W}_{i-1} - 2\bm \Lambda_i \bm \Omega_i \mathcal{\bm W}_{i-1} + \bm \Omega_i^2 \bm 1_M  \right) \right\|_\infty\\
&\leq \left\|  \bm C^T \right\|_\infty \left\| \mathcal{\bm W}_{i-1} - 2\bm \Lambda_i \bm \Omega_i \mathcal{\bm W}_{i-1} + \bm \Omega_i^2 \bm 1_M  \right\|_\infty \\
&\stackrel{(a)}{=} \left\| \mathcal{\bm W}_{i-1} - 2\bm \Lambda_i \bm \Omega_i \mathcal{\bm W}_{i-1} + \bm \Omega_i^2 \bm 1_M\right\|_\infty \\
\end{aligned}
\end{array}
\end{equation}
where the equality (a) uses the fact that $\|\bm C^T\|_\infty=1$ in that the summation of each column of $\bm C$ is 1. Since $\bm \Lambda_i$ and~$\bm \Omega_i$ are diagonal matrices with positive entries, \eqref{034} can be equivalently expressed as \cite{chen2012diffusion}:
\begin{equation}
\label{035}
\begin{array}{rcl}
\begin{aligned}
&E\left\lbrace \lVert \widetilde{\bm w}_{k,i} \rVert_2^2 \right\rbrace \leq E\left\lbrace \lVert \widetilde{\bm w}_{k,i-1} \rVert_2^2 \right\rbrace\\
&-2\chi_k \varpi_{k,i}\sqrt{\frac{E\left\lbrace \zeta_k(i)\right\rbrace - \beta E\left\lbrace \xi_k(i-1)\right\rbrace }{1-\beta}} E\left\lbrace \lVert \widetilde{\bm w}_{k,i-1} \rVert_2^2\right\rbrace \\
&+\frac{E\left\lbrace \zeta_k(i)\right\rbrace - \beta E\left\lbrace \xi_k(i-1)\right\rbrace }{1-\beta}
\end{aligned}
\end{array}
\end{equation}
for $k=1,...,N$. When the algorithm reaches the steady-state, i.e., $E\left\lbrace \lVert \widetilde{\bm w}_{k,i} \rVert_2^2 \right\rbrace = E\left\lbrace \lVert \widetilde{\bm w}_{k,i-1} \rVert_2^2 \right\rbrace$ as $i \rightarrow \infty$, from \eqref{035} we will get:
\begin{equation}
\begin{array}{rcl}
\begin{aligned}
\label{036}
&2\chi_k \varpi_{k,i} \lim \limits_{i\rightarrow \infty} E\left\lbrace \lVert \widetilde{\bm w}_{k,i-1} \rVert_2^2 \right\rbrace  \leq \\
&\;\;\;\;\;\;\;\;\;\;\;\sqrt{\lim \limits_{i\rightarrow \infty} \frac{E\{\zeta_k(i)\}-\beta E\{\xi_k(i-1)\} }{1-\beta} }.
\end{aligned}
\end{array}
\end{equation}
In view of the result that $E\{\zeta_k(i)\}$ and $E\{\xi_k(i)\}$ converge approximately to 0 as $i \rightarrow \infty$ (see Appendix B) as well as $\chi_k \neq 0 $ and $\varpi_{k,i} \neq 0$, thus, from \eqref{036} we are able to deduce that
\begin{equation}
\begin{array}{rcl}
\begin{aligned}
\label{037}
E\{\|\widetilde{\bm w}_k(\infty) \|_2^2\} \approx 0,\;\text{for}\;k=1,...,N.
\end{aligned}
\end{array}
\end{equation}
As a result, \eqref{037} illustrates that based on given assumptions, the R-dRLS algorithm can converge to the true parameter vector in the mean-square sense after enough iterations even in impulsive noise environments.

\subsection{Analysis of Evolution Behavior}
The result \eqref{037} is qualitative so that it does not predict the steady-state performance of the algorithm, due mainly to the use of the upper bound relation \eqref{020}. In this subsection, we will establish a recursive model to describe the evolution behavior of the algorithm in impulsive noise. We start by defining the following network vectors collected from all nodes:
\begin{equation}
\begin{array}{rcl}
\begin{aligned}
\label{038}
\widetilde{\bm \psi}_i&\triangleq\text{col} \{ \widetilde{\bm \psi}_{1,i}, ..., \widetilde{\bm \psi}_{N,i} \}, \\
\widetilde{\bm w}_i&\triangleq\text{col} \{ \widetilde{\bm w}_{1,i}, ..., \widetilde{\bm w}_{N,i} \},\\
\bm \varXi_i&\triangleq\text{col} \{ \bm \varXi_{1,i}, ..., \bm \varXi_{N,i} \}, \\
\end{aligned}
\end{array}
\end{equation}
where
\begin{equation}
\begin{array}{rcl}
\begin{aligned}
\label{039}
\bm \varXi_{k,i}=\sqrt{\frac{\zeta_k(i)-\beta\xi_k(i-1)}{1-\beta}} \frac{\bm g_{k,i}}{\lVert \bm g_{k,i} \rVert_2} \text{sign}(e_k(i))
\end{aligned}
\end{array}
\end{equation}
for nodes $k=1,...,N$. By these defined vectors, we can associate \eqref{017} with \eqref{018} at all the nodes:
\begin{equation}
\label{040}
\begin{array}{rcl}
\begin{aligned}
\widetilde{\bm w}_{i} = \mathcal{C}^T [\widetilde{\bm w}_{i-1} -\bm \varXi_i],
\end{aligned}
\end{array}
\end{equation}
where $\mathcal{C}= \bm C \otimes I_M$. Post-multiplying \eqref{040} by its transpose and taking the expectation, we have
\begin{equation}
\label{041}
\begin{array}{rcl}
\begin{aligned}
\bm W_{i} = &\mathcal{C}^T \left[ \bm W_{i-1} - \underbrace{E\{\widetilde{\bm w}_{i-1}\bm \varXi_i^T\}} \limits_\text{I} -\right.  \\
&\left. \underbrace{ E\{\bm \varXi_i \widetilde{\bm w}_{i-1}^T\}} \limits_\text{II} + \underbrace{E\{\bm \varXi_i \bm \varXi_i^T\} } \limits_{\text{III}} \right]  \mathcal{C},
\end{aligned}
\end{array}
\end{equation}
where $\bm W_{i} \triangleq E\{\widetilde{\bm w}_{i} \widetilde{\bm w}_{i}^T\}$ denotes the covariance matrix of the deviation vector $\widetilde{\bm w}_{i}$, and its $k$-th diagonal block of size $M\times M$, i.e., $\bm W_{k, i} \triangleq E\{\widetilde{\bm w}_{k,i} \widetilde{\bm w}_{k,i}^T\}$, represents the covariance matrix of the deviation vector $\widetilde{\bm w}_{k,i}$ at node $k$.

To evaluate terms I-III in \eqref{041}, in addition to the spatially independence in assumption 1, we also require the input regressors $\bm u_{k,i}$ to be statistically independent in time, which is also often used in analysis of distributed estimation algorithms \cite{sayed2014adaptation,sayed2014adaptive}. Therefore, performing similar manipulations as in Section IV-A on the expectations under assumptions 2-4, Lemma and Price's theorem, we can compute these three terms. Specifically, the term I in~\eqref{041} becomes
\begin{equation}
\label{042}
\begin{array}{rcl}
\begin{aligned}
E\{\widetilde{\bm w}_{i-1}\bm \varXi_i^T\} &= E\{E\{\widetilde{\bm w}_{i-1}\bm \varXi_i^T |\widetilde{\bm w}_{i-1}\} \} \\
&= \bm W_{i-1} [(\bm \Lambda_i \bm \Omega_i) \otimes \bm I_M],
\end{aligned}
\end{array}
\end{equation}
where we rewrite $\varpi_{k,i}$ contained in $\bm \Lambda_i$ as
\begin{equation}
\begin{array}{rcl}
\begin{aligned}
\label{043}
\varpi_{k,i} = \sqrt{\frac{2}{\pi}} &\left\lbrace \frac{p_{r,k}}{\sqrt{\text{Tr}\{\bm W_{k,i-1} \bm R_k\}+(\hbar+1)\sigma_{\theta}^2\}}} \right.  \\
&+\left. \frac{1-p_{r,k}}{\sqrt{\text{Tr}\{\bm W_{k,i-1} \bm R_k\}+\sigma_{\theta}^2\}}} \right\rbrace.
\end{aligned}
\end{array}
\end{equation}
The term II in \eqref{041} is the transpose of \eqref{043}. For any $k$ and $m$ belonging to the set $\{1,...,N\}$, we define the $(m,k)$-th $M\times M$ matrix $E\{\bm \varXi_i \bm \varXi_i^T\}$ as follows:
\begin{equation}
\begin{array}{rcl}
\begin{aligned}
\label{044}
E\{\bm \varXi_i \bm \varXi_i^T\}_{m,k} = E\{\bm \varXi_{m,i} \bm \varXi_{k,i}^T\}.
\end{aligned}
\end{array}
\end{equation}
When $k=m$, \eqref{044} represents the $k$-th diagonal block of $E\{\bm \varXi_i \bm \varXi_i^T\}$, which is described as
\begin{equation}
\begin{array}{rcl}
\begin{aligned}
\label{045}
E\{\bm \varXi_i \bm \varXi_i^T\}_{k,k} = \Omega_{k,i}^2 E\left\lbrace \frac{\bm R_k^{-1} \bm u_{k,i} \bm u_{k,i}^T \bm R_{k,i}^{-1}}{\sqrt{\bm u_{k,i}^T \bm R_k^{-2} \bm u_{k,i}}} \right\rbrace,
\end{aligned}
\end{array}
\end{equation}
where $\Omega_{k,i}$ is the $k$-th element of $\bm \Omega_{i}$. When $k\neq m$, the off-diagonal blocks will be simplified as
\begin{equation}
\begin{array}{rcl}
\begin{aligned}
\label{046}
&E\{\bm \varXi_i \bm \varXi_i^T\}_{m,k} = E\{E\{\bm \varXi_{m,i} \bm \varXi_{k,i}^T|\widetilde{\bm w}_{k,i-1}, \widetilde{\bm w}_{k,i-1}\}\} \\
&\simeq E\{E\{\bm \varXi_{m,i}|\widetilde{\bm w}_{m,i-1}\} \cdot E\{\bm \varXi_{k,i}^T|\widetilde{\bm w}_{k,i-1}\}\}\\
&=\chi_m\varpi_{m,i}\Omega_{m,i}E\{\widetilde{\bm w}_{m,i-1}\widetilde{\bm w}_{k,i-1}^T \}\chi_k\varpi_{k,i}\Omega_{k,i}.
\end{aligned}
\end{array}
\end{equation}
From \eqref{045} and \eqref{046}, we obtain the term III in \eqref{041}:
\begin{equation}
\begin{array}{rcl}
\begin{aligned}
\label{047}
E\{\bm \varXi_i \bm \varXi_i^T\} =&  [(\bm \Lambda_i \bm \Omega_i) \otimes \bm I_M] [\bm W_{i-1} - \breve{\bm W}_{i-1}] \times \\
&[(\bm \Lambda_i \bm \Omega_i) \otimes \bm I_M] + \breve{\bm R},
\end{aligned}
\end{array}
\end{equation}
where
\begin{equation}
\begin{array}{rcl}
\begin{aligned}
\label{048}
\breve{\bm W}_{i-1} &= \text{diag} \{\bm W_{1,i-1},...,\bm W_{N,i-1}\},\\
\breve{\bm R} &= \text{diag} \left\lbrace  \Omega_{1,i}^2 E\left\lbrace \frac{\bm R_1^{-1} \bm u_{1,i} \bm u_{1,i}^T \bm R_{1,i}^{-1}}{\sqrt{\bm u_{1,i}^T \bm R_1^{-2} \bm u_{1,i}}} \right\rbrace,...,\right. \\
&\;\;\;\;\;\;\;\;\;\;\;\;\;\;\left. \Omega_{N,i}^2 E\left\lbrace \frac{\bm R_N^{-1} \bm u_{N,i} \bm u_{N,i}^T \bm R_{N,i}^{-1}}{\sqrt{\bm u_{N,i}^T \bm R_N^{-2} \bm u_{N,i}}} \right\rbrace \right\rbrace .
\end{aligned}
\end{array}
\end{equation}
By substituting \eqref{042} and \eqref{047} into \eqref{041}, we obtain the recursive expression for $\bm W_i$:
\begin{equation}
\begin{array}{rcl}
\begin{aligned}
\label{049}
\bm W_{i} = &\mathcal{C}^T \left\lbrace \bm W_{i-1} - \bm W_{i-1} [(\bm \Lambda_i \bm \Omega_i) \otimes \bm I_M] - \right. \\
&[(\bm \Lambda_i \bm \Omega_i) \otimes \bm I_M]^T\bm W_{i-1}^T  + [(\bm \Lambda_i \bm \Omega_i) \otimes \bm I_M] \times \\
&\left. [\bm W_{i-1} - \breve{\bm W}_{i-1}] [(\bm \Lambda_i \bm \Omega_i) \otimes \bm I_M] + \breve{\bm R} \right\rbrace \mathcal{C}.
\end{aligned}
\end{array}
\end{equation}

The mean square deviation (MSD) at node $k$ is defined as $\text{MSD}_k(i) \triangleq \text{Tr} \{\bm W_{k,i}\}$, and the network MSD over all the nodes is defined as $\text{MSD}_\text{net}(i)=\frac{1}{N}\sum_{k=1}^N \text{MSD}_k(i) = \text{Tr} \{\bm W_{i}\}/N$ \cite{sayed2014adaptive}. Equation (49) models the MSD evolution behavior of the algorithm. It needs to be mentioned that to implement the model (49), $E\{\xi_k\}$ and $E\{\zeta_k\}$ defined in $\bm \Omega_i$ still need to be evaluated further. However, as shown in~\eqref{014}, $\xi_k$ and $\zeta_k$ between interconnected nodes are affected by each other and there is a comparison operation, so it is difficult to provide an evolution expression for them. In this paper, we suggest that $E\{\xi_k\}$ and $E\{\zeta_k\}$ are obtained by the ensemble average using simulations. Consequently, although (49) is a semi-analytic result, it can also be used to evaluate the convergence of the proposed algorithm.

\section{DCD-Based Algorithms}
In this section, we review the DCD-dRLS algorithm from~\cite{arablouei2013reduced}, and then develop a robust DCD-dRLS algorithm.
\subsection{The Original DCD-dRLS Algorithm}
Since the dRLS algorithm involves the matrix operation of size $M \times M$ in the computations of $\bm g_{k,i}$ and~\eqref{007} at every node, it requires a computational complexity that scales as a quadratic function of $M$ in terms of additions and multiplications per iteration $i$. To reduce the complexity, the adaptation step of the DCD-dRLS algorithm is described as~\cite{arablouei2013reduced}:
\begin{equation}
\setcounter{equation}{50}
\label{038V}
\begin{array}{rcl}
\begin{aligned}
\bm \psi_{k,i} = \bm w_{k,i-1} + \Delta \bm w_{k,i},
\end{aligned}
\end{array}
\end{equation}
where the increment $\Delta \bm w_{k,i}$ is obtained by solving the normal equation:
\begin{equation}
\label{039V}
\begin{array}{rcl}
\begin{aligned}
\bm \Phi_{k,i} \Delta\bm w_{k,i} =  \bm b_{k,i},
\end{aligned}
\end{array}
\end{equation}
\begin{equation}
\label{040V}
\begin{array}{rcl}
\begin{aligned}
\bm b_{k,i} = \lambda \bm r_{k,i-1} + e_k(i)\bm u_{k,i},
\end{aligned}
\end{array}
\end{equation}
$\bm r_{k,i-1}$ defines the residual vector at node $k$ at time instant~$i-1$:
\begin{equation}
\label{041V}
\begin{array}{rcl}
\begin{aligned}
\bm r_{k,i-1} = \bm b_{k,i-1} - \bm \Phi_{k,i-1} \Delta \hat{\bm w}_{k,i-1}.
\end{aligned}
\end{array}
\end{equation}

For reducing the complexity of computing $\Delta \hat{\bm w}_{k,i}$ and $\bm r_{k,i}$, the DCD method presented in Table \ref{table_2} is used; see \cite{zakharov2004multiplication,zakharov2008low,liu2009architecture} for details. In Table \ref{table_2}, $[\bm r_{k,i}]_l$ is the $l$-th entry of a vector $\bm r_{k,i}$, and $[\bm \Phi_{k,i}]_{l,l}$ and $[\bm \Phi_{k,i}]_{:,l}$ are the $(l,l)$-th entry and the $l$-th column of $\bm \Phi_{k,i}$, respectively.

The accuracy and complexity of the DCD method are dependent on three parameters: $H$, $M_b$, and $N_u$. In general, $H$ is chosen as a power-of-two number; $M_b$ is the number of bits being enough for a fixed-point representation of $\Delta \hat{\bm w}_{k,i}$ within an amplitude range $[-H, H]$; and $N_u$ defines a maximum number of elements in $\Delta \hat{\bm w}_{k,i}$ that can be updated at a time instant.
The DCD method only requires $2N_uM + M_b$ additions at most at each time instant with no multiplication~\cite{zakharov2008low}. Also, a larger $N_u$ makes the solution $\Delta \hat{\bm w}_{k,i}$ closer to the optimal solution $\Delta \bm w_{k,i}$ in \eqref{039V}, but increases the number of additions. It follows that if $N_u<M$, the DCD-based algorithm implements a selective partial update~\cite{dogancay2008partial}.

Similar to the dRLS algorithm, however, the DCD-dRLS algorithm will also encounter the performance deterioration when impulsive noise happens.
\begin{table}[tbp]
    \scriptsize
    \centering
    \caption{ DCD Method for Solving \eqref{039V}.}
    \label{table_2}
    \begin{tabular}{lc}
        \hline
        \text{Parameters:}$H,\;N_u,\;M_b$,\\
        \text{Initialization:} $\Delta \hat{\bm w}_{k,i} = \bm 0,\;\bm r_{k,i}= \bm b_{k,i},\;y=1,\;\mu=H/2$\\
        \hline
        \text{for} $j=1,...,N_u$\\
        \;\;\;$l= \arg \max \limits_{j=1,...,M} \{ \rvert [\bm r_{k,i}]_j \rvert \}$\\
        \;\;\;\text{while} $\rvert [\bm r_{k,i}]_l \rvert \leq (\mu/2) [\bm \Phi_{k,i}]_{l,l}$ \text{and} $y \leq M_b$\\
        \;\;\;\;\;$y=y+1$,\;$\mu=\mu/2$ \\
        \;\;\;\text{end}\\
        \;\;\;\text{if} $y > M_b$\\
        \;\;\;\;\; \text{break}\\
        \;\;\;\text{else}\\
        \;\;\;\;\; $[\Delta \hat{\bm w}_{k,i}]_l \leftarrow [\Delta \hat{\bm w}_{k,i}]_l +\mu \text{sign}([\bm r_{k,i}]_l)$ \\
        \;\;\;\;\; $\bm r_{k,i} \leftarrow \bm r_{k,i} -\mu \text{sign}([\bm r_{k,i}]_l) [\bm \Phi_{k,i}]_{:,l}$  \\
        \;\;\;\text{end}\\
        \text{end}\\
        \hline
    \end{tabular}
\end{table}
\subsection{Proposed DCD-R-dRLS Algorithm}
To achieve robustness against impulsive noise, we present here the DCD-R-dRLS algorithm.

\emph{Step 1:} At every node $k$, we firstly use the DCD method to solve the normal equation \eqref{039V} with \eqref{004} and~\eqref{040V}, yielding a solution $\Delta \hat{\bm w}_{k,i}^{(1)}$ and residual vector $\bm r_{k,i}^{(1)}$. In the presence of impulsive noise, we also impose a constraint similar to that in~\eqref{010}:
\begin{equation}
\label{042V}
\begin{array}{rcl}
\begin{aligned}
\lVert \Delta \hat{\bm w}_{k,i} \rVert_2^2 \leq \xi_k(i-1).
\end{aligned}
\end{array}
\end{equation}

\emph{Step 2:} If $\lVert \Delta \hat{\bm w}_{k,i}^{(1)} \rVert_2^2 \leq \xi_k(i-1)$, we set $\Delta \hat{\bm w}_{k,i} = \Delta \hat{\bm w}_{k,i}^{(1)}$ and $\bm r_{k,i} =  \bm r_{k,i}^{(1)}$ and then perform the update \eqref{038V}. Otherwise, we need to recalculate $\bm b_{k,i}$ in \eqref{039V} as:
\begin{equation}
\label{043V}
\begin{array}{rcl}
\begin{aligned}
\bm b_{k,i} = \lambda \bm r_{k,i-1} + \frac{\sqrt{\xi_k(i-1)}}{\lVert \Delta \hat{\bm w}_{k,i}^{(1)} \rVert_2} e_k(i)\bm u_{k,i}.
\end{aligned}
\end{array}
\end{equation}
Subsequently, based on the DCD method, we obtain the solution $\Delta \hat{\bm w}_{k,i}^{(2)}$ and the residual vector $\bm r_{k,i}^{(2)}$ from the normal equation \eqref{039V} under \eqref{004} and \eqref{043V}, thereby performing the update \eqref{038V} with the increment
\begin{equation}
\label{044V}
\begin{array}{rcl}
\begin{aligned}
\Delta \hat{\bm w}_{k,i} = \frac{\sqrt{\xi_k(i-1)}}{\lVert \Delta\hat{\bm w}_{k,i}^{(2)} \rVert_2} \Delta \hat{\bm w}_{k,i}^{(2)},
\end{aligned}
\end{array}
\end{equation}
and $\bm r_{k,i} = \bm r_{k,i}^{(2)}$.

\emph{Step 3:} The combination step \eqref{012} is performed.

\emph{Step 4:} The bound parameter $\xi_k(i)$ in the DCD-R-dRLS algorithm is updated according to
\begin{equation}
\label{045V}
\begin{array}{rcl}
\begin{aligned}
\zeta_k(i) =& \beta  \xi_k(i-1) + (1-\beta) \lVert \Delta\hat{\bm w}_{k,i} \rVert_2^2, \\
\xi_k(i) =&\sum \limits_{m\in\mathcal{N}_k}c_{m,k} \zeta_m(i).
\end{aligned}
\end{array}
\end{equation}

Table \ref{table_3} summarizes the DCD-R-dRLS algorithm.
\begin{table}[tbp]
    \scriptsize
    \centering
    \caption{ Proposed DCD-R-$\rm d$RLS Algorithm}
    \label{table_3}
    \begin{tabular}{lc}
        \hline
        \text{Parameters:} $0\ll \beta <1$, $\lambda$, $\delta$ and $E_c$\\
        \text{Initialization}: $\bm w_{k,0} = \bm 0$, $\bm \Phi_{k,0}=\delta \bm I_M$ and $\xi_k(0)= E_c \frac{\sigma_{d,k}^2}{M\sigma_{u,k}^2}$\\
        \hline
        \text{for each node} $k$:\\
        $e_k(i) = d_k(i)-\bm u_{k,i}^T \bm w_{k,i-1}$ \\
        $\bm \Phi_{k,i} = \lambda \bm \Phi_{k,i-1} + \bm u_{k,i}\bm u_{k,i}^T$ \\
        $\bm b_{k,i} = \lambda \bm r_{k,i-1} + e_k(i)\bm u_{k,i}$\\
        \text{Using DCD to solve} $\bm \Phi_{k,i} \Delta\bm w_{k,i} = \bm b_{k,i}$, \text{yielding}\\
        $\Delta \hat{\bm w}_{k,i} = \Delta \hat{\bm w}_{k,i}^{(1)}$ and $\bm r_{k,i} = \bm r_{k,i}^{(1)}$\\
        \text{if} $\lVert \Delta \hat{\bm w}_{k,i}\rVert_2^2 > \xi_k(i-1)$ \\
        \;\;\;\;$\bm b_{k,i} = \lambda \bm r_{k,i-1} + \frac{\sqrt{\xi_k(i-1)}}{\lVert \Delta \hat{\bm w}_{k,i} \rVert_2} e_k(i)\bm u_{k,i}$\\
        \;\;\;\;\text{Using DCD to solve} $\bm \Phi_{k,i} \Delta\bm w_{k,i} = \bm b_{k,i}$, \text{yielding}\\
        \;\;\;\;$\Delta \hat{\bm w}_{k,i} = \frac{\sqrt{\xi_k(i-1)}}{\lVert \Delta\hat{\bm w}_{k,i}^{(2)} \rVert_2} \Delta \hat{\bm w}_{k,i}^{(2)}$ and $\bm r_{k,i} = \bm r_{k,i}^{(2)}$\\
        \text{end}\\
        $\bm \psi_{k,i} = \bm w_{k,i-1} + \Delta \hat{\bm w}_{k,i}$\\
        $\bm w_{k,i} = \sum\limits_{m\in\mathcal{N}_k}c_{m,k} \bm \psi_{m,i}$\\
        $\zeta_k(i) = \beta \xi_k(i-1) + (1-\beta) \lVert \Delta\hat{\bm w}_{k,i} \rVert_2^2$\\
        $\xi_k(i) =\sum\limits_{m\in\mathcal{N}_k}c_{m,k} \zeta_m(i)$\\
        \hline
    \end{tabular}
\end{table}

\textbf{Remark 3:} An impulsive noise sample appearing at time instant $i$ would yield a mismatch solution $\Delta\hat{\bm w}_{k,i}^{(1)}$ so that $\left. \lVert \Delta\hat{\bm w}_{k,i}^{(1)} \rVert_2^2 > \xi_k(i-1) \right.$. In this case, the scaling factor $\frac{\sqrt{\xi_k(i-1)}}{\lVert \Delta\hat{\bm w}_{k,i}^{(1)} \rVert_2}$ in \eqref{043V} is small enough to eliminate impulsive noise hidden in $e_k(i)$. A similar scaling factor $\frac{\sqrt{\xi_k(i-1)}}{\lVert \Delta\hat{\bm w}_{k,i}^{(2)} \rVert_2}$ in \eqref{044V} is to make the increment satisfy the constraint \eqref{042V}. Consequently, the DCD-R-dRLS algorithm improves the robustness to impulsive noise relative to the DCD-dRLS algorithm. Moreover, the decreasing sequence $\{\xi_k(i)\}$ shown in \eqref{045V} further guarantees the robustness. It is worth noting that due to $\rVert \bm g_{k,i} \lVert_2 \lvert e_k(i)\rvert \approx \lVert \Delta\hat{\bm w}_{k,i} \rVert_2$, the DCD-R-dRLS algorithm is a DCD-based variant of the R-dRLS algorithm. Unlike the R-dRLS algorithm, based on the NC method we re-initialize $\bm \Phi_{k,i}$ with $\bm \Phi_{k,0}$ to endow the DCD-R-dRLS algorithm with the tracking capability when $\bm w^o$ suddenly changes.

\textbf{Remark 4:} Let $C_{dcd}^+$ denote the only required number of additions for the DCD algorithm, with $C_{dcd}^+ \leq 2N_uM + M_b$. In Table \ref{table_4}, we provide the computational complexity of the existing dLMS, dRLS, DCD-dRLS, and both proposed R-dRLS and DCD-R-dRLS algorithms at node \emph{k} per time instant~$i$, where $n_k$ denotes the cardinality of $\mathcal{N}_k$. For shift structured input regressor at node $k$~\cite{chouvardas2011adaptive, li2010distributed}, i.e., $\bm u_{k,i}=[u_k(i),u_k(i-1),...,u_k(i-M+1)]^T$, where $u_k(i)$ is an input sample at time instant $i$, implementing $\bm \Phi_{k,i}$ in \eqref{004} is very simplified. In this situation, by copying the upper-left $(M-1)\times (M-1)$ block of $\bm \Phi_{k,i-1}$ leads to the lower-right $(M-1)\times (M-1)$ block of $\bm \Phi_{k,i}$. The remaining part of $\bm \Phi_{k,i}$ that needs to be updated is the first row and first column. Owing to symmetry of $\bm \Phi_{k,i}$, only calculating the first column is sufficient, which is formulated as:
\begin{equation*}
\begin{array}{rcl}
[\bm \Phi_{k,i}]_{:,1} = \lambda [\bm \Phi_{k,i-1}]_{:,1} + u_k(i)\bm u_{k,i}.
\end{array}
\end{equation*}
Note that, in the DCD-R-dRLS algorithm, $\left. \kappa=1\right. $ represents the case $\left. \lVert \Delta \hat{\bm w}_{k,i}\rVert_2^2 > \xi_k(i-1)\right. $ at time instant $i$ (which leads to the maximum complexity), otherwise $\kappa=0$. The comparisons required in the R-dRLS and DCD-R-dRLS algorithms are counted as additions.

\begin{figure}[htb]
\centering
\includegraphics[scale=0.53] {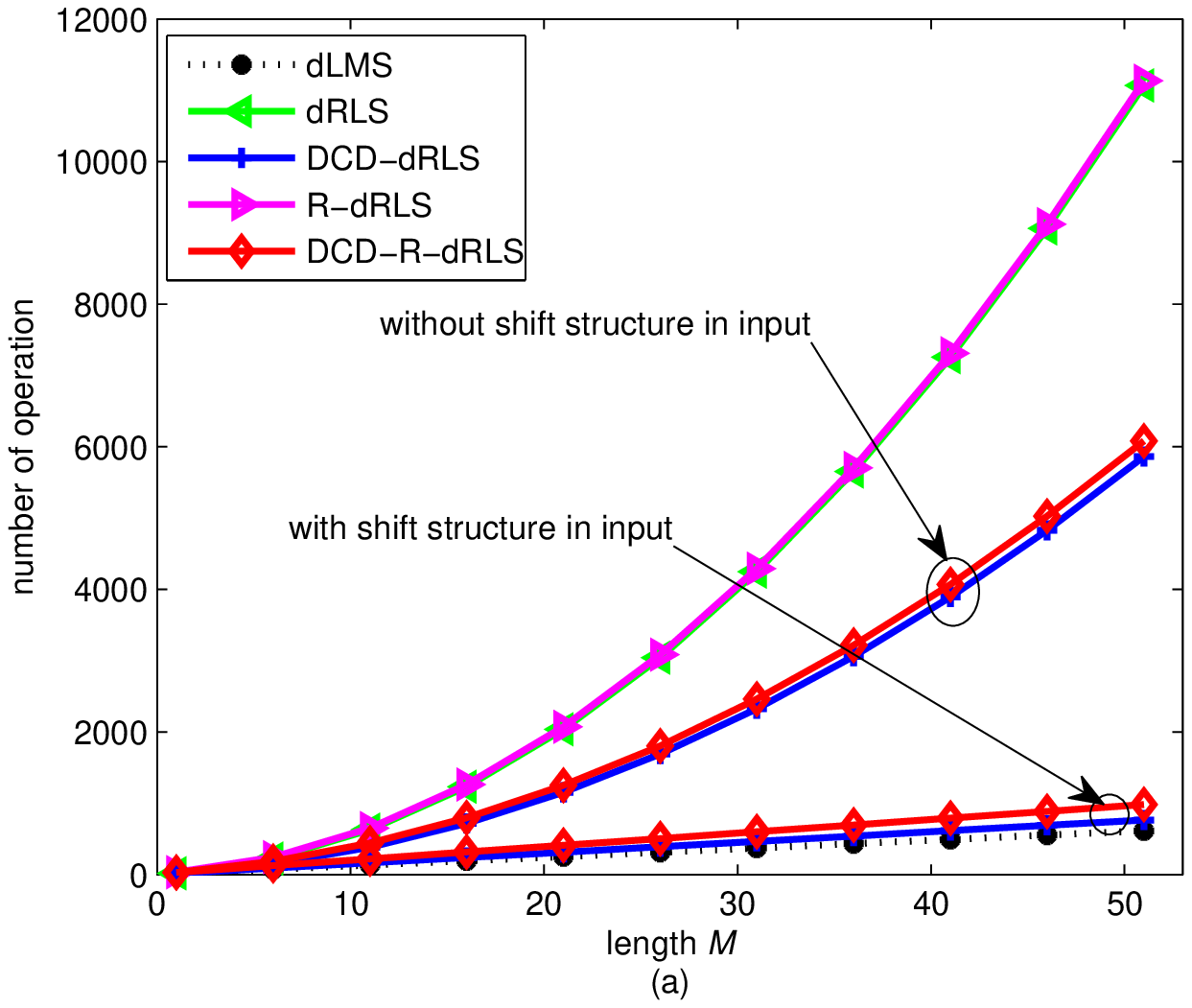}
\centering
\includegraphics[scale=0.53] {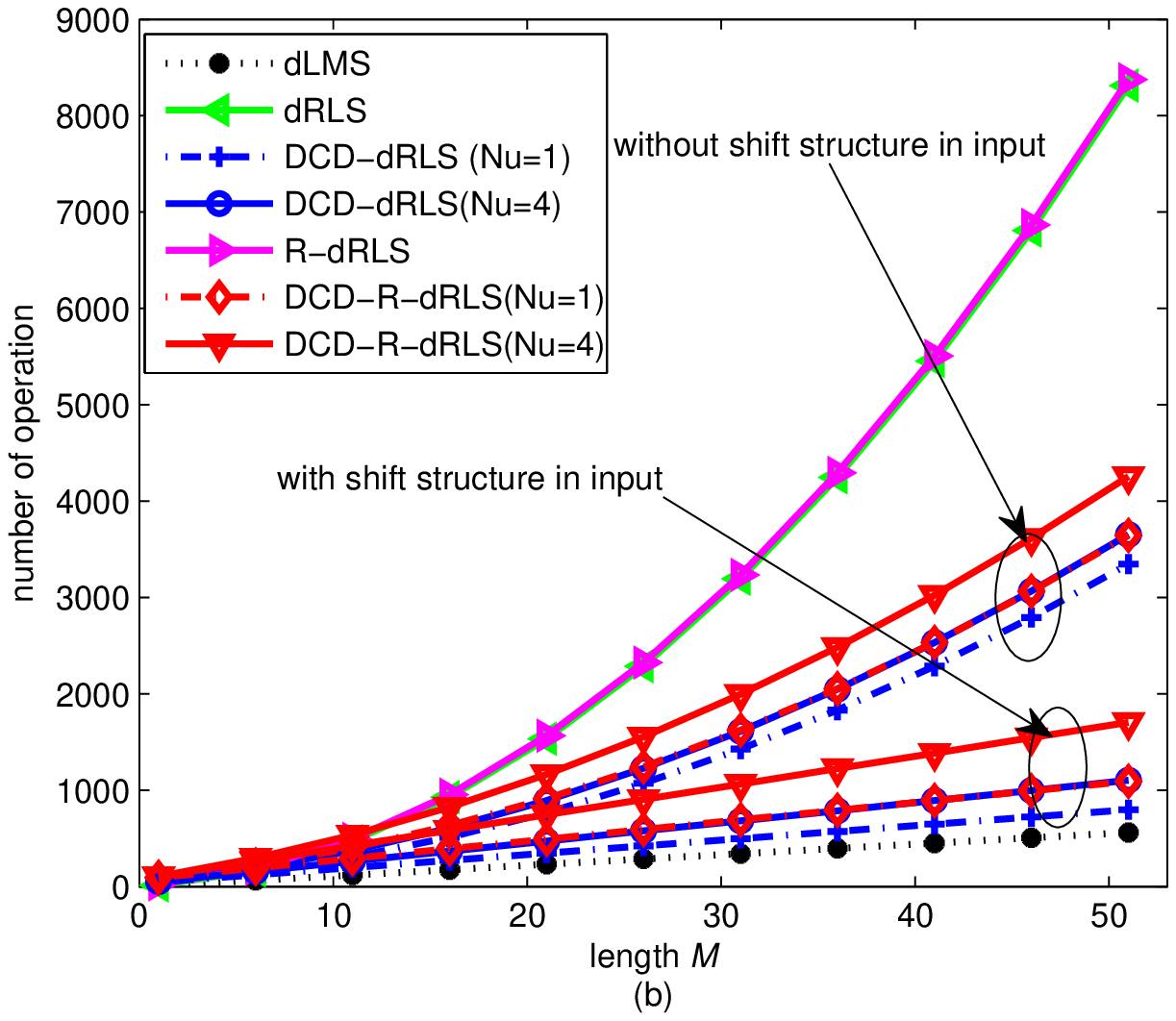}
\caption{Complexity of the algorithms versus the length of the target vector at node $k$. (a) multiplications and (b) additions. }
\label{Fig2}
\end{figure}

Consider an example with $n_k =10$, $M_b=32$ and $\kappa=1$, Fig. \ref{Fig2} depicts the number of operations of some diffusion algorithms in terms of multiplications and additions at node $k$ at each time instant versus $M$. It is clear that the computational complexity of the dLMS algorithm, with the order of $\mathcal{O}(M)$, is much lower than that of the dRLS algorithm. As expected, since $N_u<M$, compared with the standard dRLS and R-dRLS algorithms, their DCD versions obtain about $50\%$ reduction in both multiplications and additions for the case of general input regressors. However, for shift structured input regressors, the computational cost is drastically reduced from the order $\mathcal{O}(M^2)$ to $\mathcal{O}(M)$, which is more pronounced in scenarios with large $M$. Moreover, the multiplications required in the DCD-based algorithms are not dependent of $N_u$. On the other hand, in contrast with the existing dRLS and DCD-dRLS algorithms, the additional complexities in the proposed R-dRLS and DCD-R-dRLS algorithms resulted from the computations of the scaling factor and the bound parameter are small. In addition to the complexity, for both proposed algorithms, each node $k$ increases communication cost of $n_k-1$ numbers for transmitting~$\zeta_k$ to its neighbors.
\begin{table*}[tbp]
    \scriptsize
    \centering
    \caption{Computational complexity of algorithms for node \emph{k} per time instant.}
    \label{table_4}
    \begin{tabular}{l|cccc}
        \hline
        \textbf{Algorithms} &\textbf{Multiplications} &\textbf{Additions} &\textbf{Divisions} &\textbf{Square-root}\\
        \hline
        \text {dLMS}     &$n_kM+2M+1$ &$n_kM+M$ &- &-  \\
        \hline
        \text {dRLS}  &$n_kM+4M^2+3M$ &$n_kM+3M^2$ &$M$ &-  \\
        \hline
        \text {DCD-dRLS}\\without shift structure in input &$n_kM+2M^2+3M$ &$n_kM+M^2+2M+C_{dcd}^+$ &- &-  \\
        \hline
        \text {DCD-dRLS}\\with shift structure in input &$n_kM+5M$ &$n_kM+3M+C_{dcd}^+$ &- &-  \\
        \hline
        \textbf {R-dRLS} &$n_k(M+1)+4M^2+4M+5$ &$n_k(M+1)+3M^2+M+1$ &$M+1$ &1  \\
        \hline
        \textbf {DCD-R-dRLS}\\without shift structure in input &$n_k(M+1)+2M^2+4M+3\kappa M+2$ &$n_k(M+1)+M^2+3M+\kappa(2M-1+C_{dcd}^+)+C_{dcd}^+$ &$2\kappa$ &$2\kappa$ \\
        \hline
        \textbf {DCD-R-dRLS} \\with shift structure in input &$n_k(M+1)+6M+3\kappa M+2$ &$n_k(M+1)+4M+\kappa(2M-1+C_{dcd}^+)+C_{dcd}^+$ &$2\kappa$ &$2\kappa$ \\
        \hline
    \end{tabular}
\end{table*}

\textbf{Remark 5:} From the DCD-R-dRLS algorithm, we can directly obtain its special form for a single-agent scenario, referred it to as the DCD-R-RLS algorithm. In other words, the DCD-R-RLS algorithm is the DCD implementation of the algorithm presented in~\cite{vega2009fast}.

\section{Simulation Results}
Simulation examples are presented for a diffusion network with $N=20$ nodes on distributed parameter estimation and distributed spectrum estimation. The network topology adopted for all simulations is shown in Fig.~\ref{Fig3}(a), unless otherwise specified. Herein, we do not consider the measurement sharing in the adaptation step for all diffusion algorithms. The Metropolis rule~\cite{takahashi2010diffusion} used for computing the combination coefficients~$\{c_{m,k}\}$ in combination step is expressed as:
\begin{equation*}
\begin{array}{rcl}
\begin{aligned}
c_{m,k}=
\setlength{\nulldelimiterspace}{0pt}
\left\{
\begin{IEEEeqnarraybox}[\relax][c]{l's}
1/\max(n_m,n_k),\;\text{if}\;m\in\mathcal{N}_k,\;m\neq k\\
1-\sum \limits_{m\neq k}c_{m,k}, \text{if}\;m=k\\
0,\;\text{otherwise}.
\end{IEEEeqnarraybox}
\right.
\end{aligned}
\end{array}
\end{equation*}
\subsection{Distributed Parameter Estimation}
\begin{figure}[htb]
    \centering
    \includegraphics[scale=0.57] {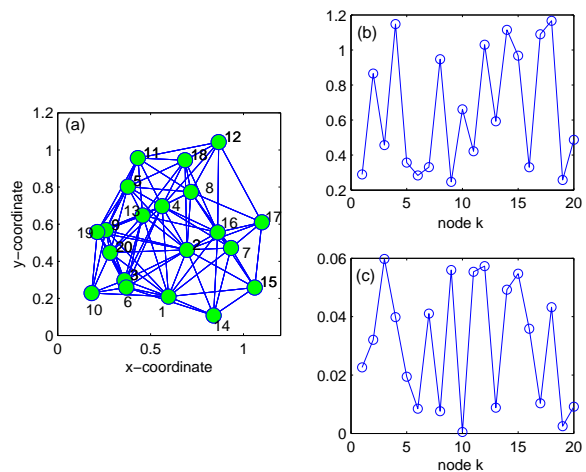}
    \hspace{2cm}\caption{(a) topology of newtwork with 20 nodes, and profiles of (b) $\sigma_{\epsilon,k}^2$ and (c) $\sigma_{\theta,k}^2$ per node $k$.}
    \label{Fig3}
\end{figure}

The vector $\bm w^o$ to be estimated has a length of $M=16$ and a unit norm; it is generated randomly from a zero-mean uniform distribution. The input regressor $\bm u_{k,i}$ has a shift structure, where $u_k(i)$ is colored and generated by a second-order autoregressive system:
\begin{equation*}
u_k(i) = 1.6u_k(i-1)-0.81u_k(i-2)+ \epsilon_k(i),
\end{equation*}
where $\epsilon_k(i)$ is a zero-mean white Gaussian process with variance $\sigma_{\epsilon,k}^2$. The background noise $\theta_k(i)$ is zero-mean white Gaussian noise with variance $\sigma_{\theta,k}^2$. Variances $\sigma_{\epsilon,k}^2$ and $\sigma_{\theta,k}^2$ are shown in Fig. \ref{Fig3}(b) and (c), respectively, for all the nodes. We employ the network MSD to assess the performance of algorithms.
All results are the average over 200 independent trials.

\begin{figure}[htb]
    \centering
    \includegraphics[scale=0.53] {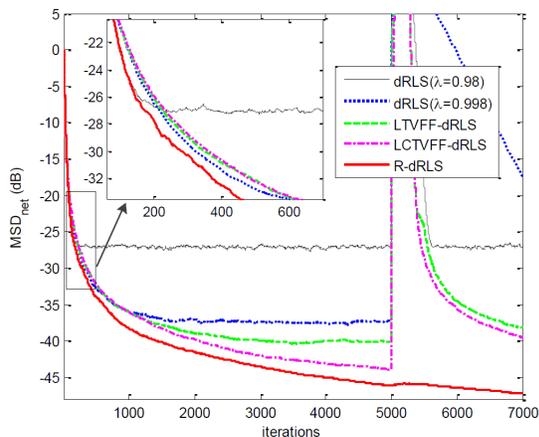}
    \hspace{2cm}\caption{Network MSD curves of the algorithms. [Gaussian noise with a cluster of impulses]. Parameter setting of the algorithms (with notations from references) is as follows: $\alpha$=0.97, $\beta$=0.0005, $\lambda_+$=0.9998 and $\lambda_-$=0.95 (LTVFF-dRLS); $\alpha$=0.8, $\beta$=0.015, $\gamma$=0.95, $\lambda_+$=0.9998 and $\lambda_-$=0.95 (LCTVFF-dRLS); $\lambda$=0.98, $\beta$=0.97 and $E_c$=10 (R-dRLS).}
    \label{Fig4}
\end{figure}

\textit{Example 1: Except for the background noise $\theta_k(i)$, a cluster of impulses with length $200$ is also added to corrupt $d_k(i)$ at iteration $i=5001$ }\footnote{Such a scenario is similar to double-talk in echo cancellation.}. The cluster is drawn from a zero-mean white Gaussian process, but with a large variance $1000\sigma_{y,k}^2$ to generate impulsive samples, where $\sigma_{y,k}^2$ denotes the power of $y_k(i) = \bm u_{k,i}^T\bm w^o$. Fig.~\ref{Fig4} compares the performance of the proposed R-dRLS algorithm with that of the dRLS and both LTVFF-dRLS and LCTVFF-dRLS algorithms presented in~\cite{zhang2017variable}. The parameters of the algorithms are set to make a comparable convergence rate. The regularization constant for all RLS-type algorithms is chosen as $\delta=0.01$. It is clear to see, for a small forgetting factor $\lambda=0.98$, the conventional dRLS algorithm converges faster but has a higher estimation error; conversely, by increasing the forgetting factor, it has a lower estimation error but its convergence rate becomes slower. In particular, using a large forgetting factor $\lambda=0.998$, the dRLS will need more time to converge again after a cluster of impulses enforces the algorithm to diverge. Due to the use of variable forgetting factor schemes, both LTVFF-dRLS and LCTVFF-dRLS algorithms solve this performance trade-off to a certain extent. As stated in Remark 2, the R-dRLS algorithm also overcomes this performance trade-off since it employs a variable 'step-size' factor in the adaptation step. Besides, unlike the dRLS, LTVFF-dRLS and LCTVFF-dRLS algorithms, even though a cluster of impulses does not happen until the algorithms reach the steady-state, the R-dRLS algorithm also does not undergo divergence. This is because the R-dRLS algorithm can judge by \eqref{011} whether impulses occur or not and perform corresponding updates.

\textit{Example 2: The additive noise $v_k(i)$ is a CG process given in assumption 4.} At every node $k$, we set $p_{r,k}$ as a random number in the range of $[0.001,0.05]$ and $\sigma_{g,k}^2=1000\sigma_{y,k}^2$. For a fair comparison of RLS-type algorithms, we choose the same forgetting factor $\lambda$=0.985 and regularization constant $\delta$=0.01, except $\delta$=0.5 in the dRLP and RVWC-dRLS algorithms.

\begin{figure}[htb]
    \centering
    \includegraphics[scale=0.53] {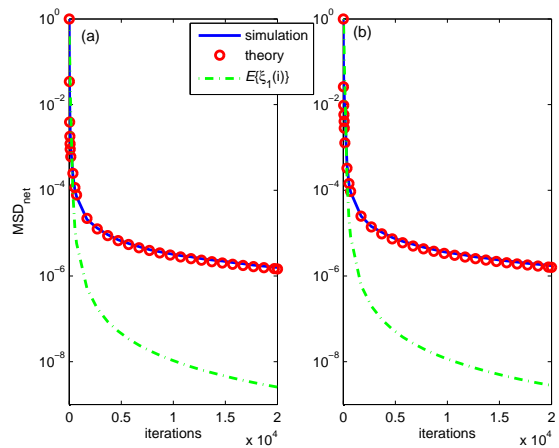}
    \hspace{2cm}\caption{ Verification of (49) for the R-dRLS algorithm (with the parameters $\beta$=0.97 and $E_c$=1). (a) $p_{r,k}=0.01$ and (b) $p_{r,k}=0.05$. }
    \label{Fig5}
\end{figure}

Fig.~\ref{Fig5} checks the validity of the semi-analytic result~(49), where we plot $E\{\xi_1(i)\}$ at node 1 (having similar results at other nodes). To take into account the assumption on input regressors $\bm u_{k,i}$ in analysis, here its entries are generated from a white Gaussian process $\epsilon_k(i)$. To compute (49), we use the same impulsive noise parameters: $p_{r,k}=0.01$ or 0.05, and $\sigma_{g,k}^2=10000 \sigma_{\theta,k}^2$ at all the nodes. As one can see, the theoretical results have good fit with the simulated results. Moreover, $E\{\xi_1(i)\}$ obtained by the ensemble average of simulations is a decreasing function of the iteration $i$, which further supports the theory in Appendix~B.

\begin{figure}[htb]
    \centering
    \includegraphics[scale=0.53] {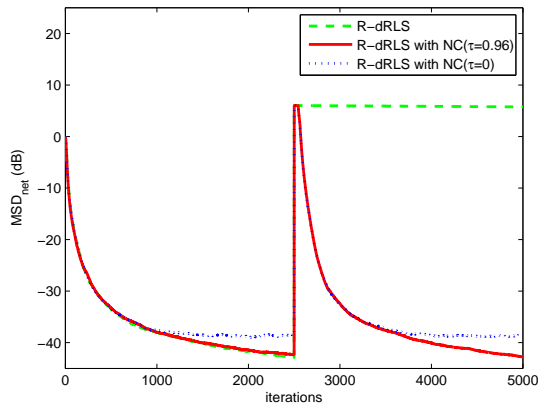}
    \hspace{2cm}\caption{ Effect of the NC method. Parameter setting of algorithms: $\beta$=0.97 and $E_c$=1 (R-dRLS); $\varrho$=3 and $t_{th}$=15 (NC).  }
    \label{Fig6}
\end{figure}

Fig.~\ref{Fig6} investigates the effect of the NC method on the R-dRLS algorithm. It can be seen that the R-dRLS algorithm will not re-converge after $\bm w^o$ changes to $-\bm w^o$ at iteration~$i=2501$. In this scenario, all algorithms have a large sharp phase transition of MSD due to the mismatch between $-\bm w^o$ and its estimate $\bm w_{k,i}$ at that moment. The NC method can endow the R-dRLS algorithm with good tracking capability for such a change of $\bm w^o$. Benefited from the smoothing operation~\eqref{015}, the NC ($\left. \tau=0.96\right. $) only slightly degrades the steady-state performance of the R-dRLS algorithm compared with the non-smooth version in \cite{YYu2018} (i.e., $\left. \tau=0\right. $).

\begin{figure}[htb]
    \centering
    \includegraphics[scale=0.53] {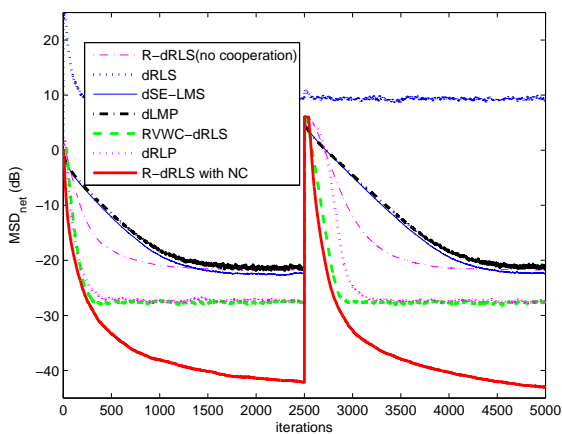}
    \hspace{2cm}\caption{Network MSD curves of the algorithms in impulsive noise with BG distribution. Parameter setting of algorithms (with notations from references) is as follows: $\mu_k$=0.015 (dSE-LMS); $\mu_k$=0.015 and $p$=1.3 (dLMP); $p$=1.3 (dRLP); $\beta$=0.97 and $E_c$=1 (R-dRLS); $\varrho$=3, $\tau$= 0.96 and $t_{th}$=15 (NC). According to Remark 1, parameters of RVWC-dRLS are chosen as $L$=16, $\alpha$=2.58 and $\lambda$=0.97 (see \cite{ahn2017new} for detailed design of RVWC).}
    \label{Fig7}
\end{figure}

In Fig.~\ref{Fig7}, we compare the performance of the dRLS, dSE-LMS, dLMP, RVWC-dRLS, and dRLP algorithms with that of the proposed R-dRLS with NC algorithm. Note that, the R-dRLS (no cooperation) is that each node performs a standalone adaptive algorithm presented in~\cite{vega2009fast}. As expected, the dRLS algorithm has a poor performance in the presence of impulsive noise, while other algorithms are robust. Among these robust algorithms, the convergence of dSE-LMS and dLMP algorithms is slow. Thanks to the decorrelation property of dRLS, the RVWC-dRLS, dRLP, and R-dRLS with NC algorithms obtain fast convergence. In particular, the proposed R-dRLS with NC algorithm has also a large reduction in the steady-state MSD. This is due mainly to the fact that its updated energy described by \eqref{010} and \eqref{014} diminishes with iterations.

\begin{figure}[htb]
    \centering
    \includegraphics[scale=0.53] {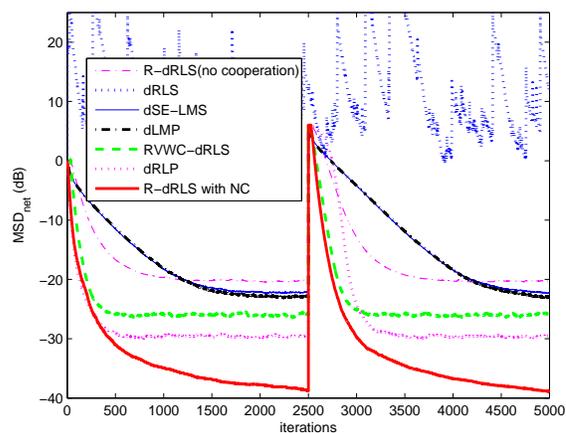}
    \hspace{2cm}\caption{Network MSD curves of algorithms. [$\alpha$-stable noise]. Parameters in some of algorithms are tuned as follows: $p=1.18$ (dLMP and dRLP); $\varrho$=2 and $t_{th}$=5 (NC).}
    \label{Fig8}
\end{figure}

\begin{figure}[htb]
    \centering
    \includegraphics[scale=0.53] {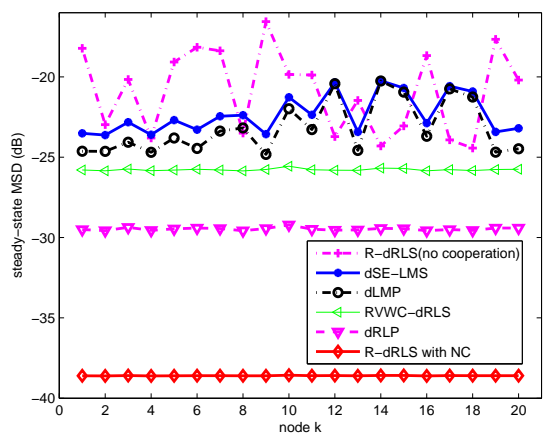}
    \hspace{2cm}\caption{Node-wise steady-state MSD of the algorithms. [$\alpha$-stable noise].}
    \label{Fig9}
\end{figure}

\textit{Example 3: The additive noise $v_k(i)$ here is generated by the $\alpha$-stable process, also called the $\alpha$-stable noise.} Its characteristic function is given by $\varphi(t)=\exp(-\gamma \lvert t \lvert^\alpha)$~\cite{pelekanakis2014adaptive,shao1993signal}, where the characteristic exponent $\alpha \in (0,2]$ describes the impulsiveness of the noise (smaller $\alpha$ leads to more impulsive noise samples) and $\gamma>0$ represents the dispersion level of the noise. In particular, when $\alpha=2$, it reduces to the Gaussian noise. It is rare to find $\alpha$-stable noise with $\alpha<1$ in practice \cite{pelekanakis2014adaptive,shao1993signal}. In this example, thus we set $\alpha=1.2$ and $\gamma=2/15$. The learning performance of the algorithms is shown in Fig.~\ref{Fig8}. Fig. \ref{Fig9} shows the node-wise steady-state MSD of the robust algorithms (i.e., excluding the dRLS), by averaging MSD values from iteration 2 400 to 2500. As can be seen from Figs.~\ref{Fig8} and~\ref{Fig9}, the proposed R-dRLS algorithm with NC outperforms the known robust diffusion algorithms in terms of convergence rate, steady-state accuracy and tracking capability. As shown in Fig.~\ref{Fig7} to Fig.~\ref{Fig9}, due to the cooperation of interconnected nodes, the R-dRLS algorithm improves the estimation performance compared with its non-cooperative counterpart.

\begin{figure}[htb]
    \centering
    \includegraphics[scale=0.4] {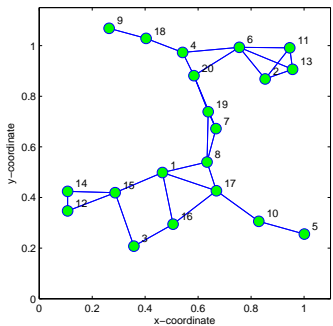}
    \hspace{2cm}\caption{Topology of a less connected network.}
    \label{Fig10}
\end{figure}

\begin{figure}[htb]
    \centering
    \includegraphics[scale=0.53] {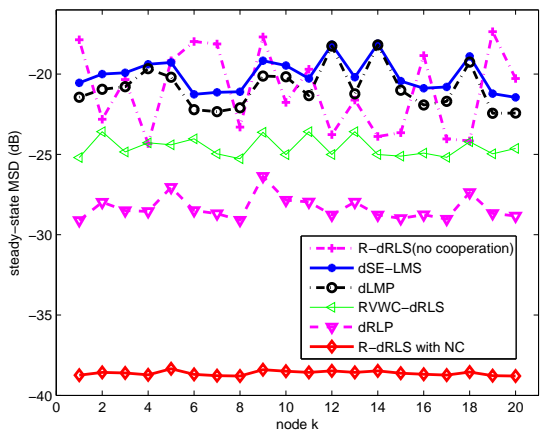}
    \hspace{2cm}\caption{Node-wise steady-state MSD of the algorithms for the network depicted in Fig.~\ref{Fig10}. [$\alpha$-stable noise]. Parameters setting of the algorithms is the same as Fig.~\ref{Fig8}.}
    \label{Fig11}
\end{figure}

We also perform the simulations for the network in Fig.~\ref{Fig10} with less connections among nodes. Fig. \ref{Fig11} shows the node-wise steady-state MSD of those algorithms in Fig. \ref{Fig9}. By comparing these two figures, it is seen that the proposed R-dRLS algorithm is more likely to reach the same estimates at all nodes.

\begin{figure}[htb]
    \centering
    \includegraphics[scale=0.53] {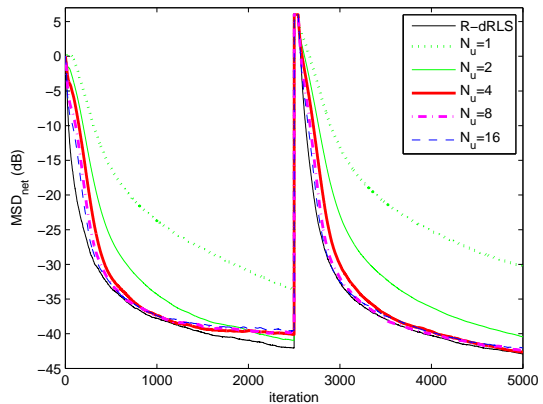}
    \hspace{2cm}\caption{Network MSD curves of the DCD-R-dRLS algorithm in CG noise. Parameters choice of the DCD-R-dRLS is the same as the R-dRLS in Fig.~\ref{Fig7} except $\lambda=0.975$, $\beta=0.96$ and $\tau=0.97$. }
    \label{Fig12}
\end{figure}

\begin{figure}[htb]
    \centering
    \includegraphics[scale=0.53] {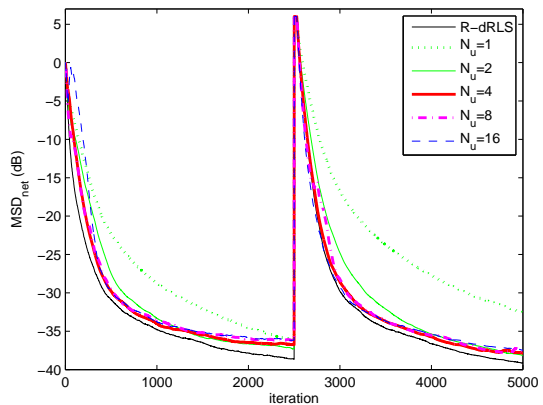}
    \hspace{2cm}\caption{Network MSD curves of the DCD-R-dRLS algorithm in $\alpha$-stable noise. Parameters choice of the DCD-R-dRLS is the same as the R-dRLS in Fig.~\ref{Fig8} except  $\lambda=0.975$, $\beta=0.96$ and $\tau=0.97$.}
    \label{Fig13}
\end{figure}

\textit{Example 4: Comparison of DCD-algorithms.} Figs. \ref{Fig12} and~\ref{Fig13} compare the DCD-R-dRLS algorithm using different $N_u$ values with its standard version in CG-noise and $\alpha$-noise scenarios\footnote{Here the curves of both the R-dRLS and DCD-dRLS algorithms are omitted due to their divergence performance in impulsive noise.}. The DCD parameters are $H=4$ and $M_b=16$. It is seen that, the proposed DCD-R-dRLS algorithm is also robust to impulsive noises, and approaches the R-dRLS performance with increase in $N_u$. In this example, the DCD-R-dRLS algorithm with $N_u=4\;(<M)$ has a good approximation to the R-dRLS algorithm, while the complexity of the former is significantly lower than that of the latter. Moreover, many simulations have been carried out in different impulsive noise scenarios by prolonging the iteration~$i$ to a larger number than the one in Fig. \ref{Fig5}, e.g., $5\times10^{5}$, using MATLAB R2013A on a Intel(R) Core(TM) i5-4590 CPU @ 3.30 GHz processor. We did not observe any numerical instability during the simulations for both proposed R-dRLS and DCD-dRLS algorithms.

\subsection{Application: Distributed Spectrum Estimation}
We have also tested the proposed algorithms' performance in an application of distributed spectrum estimation in CR, in which the objective is to estimate the spectrum of a transmitted signal source $s$ in the network with $N$ nodes \cite{di2013distributed,miller2016distributed,zhang2017variable}. We use $\bm \phi_s(f)=\sum_{m=1}^{M} q_m(f) w_m^o=\bm q^T(f) \bm w^o$ to denote the power spectral density (PSD) of the signal $s$ at frequency~$f$, where $\bm q(f)=[q_1(f),...,q_M(f)]^T$ is a vector consisting of basis functions evaluated at normalized frequency $f$, and $\bm w^o=[w_1^o,...,w_M^o]^T$ stands for the power that transmits the signal $s$ over each of $M$ basis functions and needs to be estimated. Such basis expansion can accurately model the spectrum of the signal $s$ for large enough $M$. Considering $\mathcal{H}_k(f,i)$ is the transfer function of the channel between the station emitting the signal $s$ and receiver node $k$ at time instant~$i$, the PSD of the received signal at node $k$ can be expressed as
\begin{equation}
\label{046V}
\begin{array}{rcl}
\begin{aligned}
\bm \phi_{k,r}(f) &= |\mathcal{H}_k(f,i)|^2 \bm \phi_s(f) + \sigma_{r,k}^2\\
&=\bm q_{k,i}^T(f) \bm w^o + \sigma_{r,k}^2,
\end{aligned}
\end{array}
\end{equation}
where $\bm q_{k,i}(f) = |\mathcal{H}_k(f,i)|\bm q(f)$, and $\sigma_{r,k}^2$ is the received noise power at node $k$.

At time instant $i$, each node $k$ observes the received PSD expressed in \eqref{046V} over $N_c$ frequency samples $f_\iota=f_{\min}:(f_{\max}-f_{\min})/N_c:f_{\max}$ for $\iota=1,...,N_c$; accordingly, the output measurements of node obey the following relation:
\begin{equation}
\label{047V}
\begin{array}{rcl}
\begin{aligned}
d_{k}^\iota(i) = \bm q_{k,i}^T(f_\iota) \bm w^o + \sigma_{r,k}^2+v_k^\iota(i),
\end{aligned}
\end{array}
\end{equation}
where $v_k^\iota(i)$ denotes the observation noise at frequency $f_\iota$. The noise power $\sigma_{r,k}^2$ can be estimated with high accuracy before the spectrum estimation, using, for example, an energy estimator over an idle band, and then subtracted from \eqref{047V}~\cite{di2013distributed,miller2016distributed,zhang2017variable}. Then, by collecting the output measurements over $N_c$ frequencies, we obtain a data model at every node $k$ for distributed spectrum estimation:
\begin{equation}
\label{048V}
\begin{array}{rcl}
\begin{aligned}
\bm d_{k}(i) = \bm Q_{k,i}\bm w^o + \bm v_k(i),
\end{aligned}
\end{array}
\end{equation}
where $\bm Q_{k,i} = [\bm q_{k,i}(f_1),...,\bm q_{k,i}(f_{N_c})]^T$, $\bm d_{k}(i)=[d_{k}^1(i),...,d_{k}^{N_c}(i)]^T$, and $\bm v_{k}(i)=[v_{k}^1(i),...,v_{k}^{N_c}(i)]^T$.

\begin{figure}[htb]
    \centering
    \includegraphics[scale=0.53] {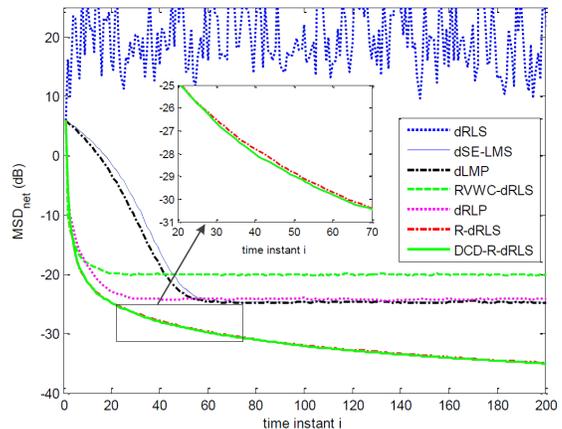}
    \hspace{2cm}\caption{Network MSD curves of various diffusion algorithms for distributed spectrum estimation. Some parameters of algorithms are re-tuned as follows: $\mu_k=0.012$ (dSE-LMS); $\mu_k=0.016$ (dLMP); $\lambda=0.997$ (dRLP, RVWC-dRLS); only $\xi_k(0)=1$ (R-dRLS, DCD-R-dRLS) differing from Fig.~\ref{Fig7}.}
    \label{Fig14}
\end{figure}

\begin{figure}[htb]
    \centering
    \includegraphics[scale=0.53] {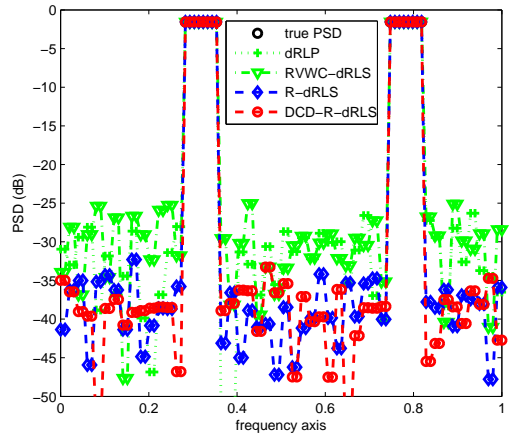}
    \hspace{2cm}\caption{PSD curves of various diffusion RLS algorithms.}
    \label{Fig15}
\end{figure}

Based on this model, we estimate the unknown spectrum $\bm w^o$ of the signal $s$ using different diffusion algorithms over the network given in Fig.~\ref{Fig3}(a). In the simulation~\cite{miller2016distributed,zhang2017variable}, we use $M=50$ nonoverlapping rectangular basis functions\footnote{Other basis functions are also possible, e.g., raised cosines, or Gaussian bells \cite{di2013distributed}.} with amplitude equal to one to model the PSD of the signal~$s$. The nodes scan $N_c=100$ frequencies over the normalized frequency axis between 0 and 1. We assume that $\bm w^o$ has only 8 non-zero elements, meaning that the unknown spectrum is transmitted over 8 basis functions, and the power transmitted over each basis function is set to 0.7. The observation noise $v_k^\iota(i)$ is an $\alpha$-stable process as in the previous Example~3~\cite{ZHU201594}. In~Fig. \ref{Fig14}, we compare the network MSD performance of different algorithms considered for the distributed spectrum estimation. As depicted, the dRLS algorithm can not identify the spectrum coefficients $\bm w^o$ due to its divergence in an $\alpha$-stable noise environment. In comparison with the dSE-LMS, dLMP, RVWC-dRLS and dRLP algorithms, the proposed R-dRLS and DCD-R-dRLS (with $N_u=4$) algorithms still obtain better estimation performance. We also notice from this figure that the DCD-R-dRLS algorithm with lower computational complexity approaches the R-dRLS performance. In Fig.~\ref{Fig15}, we also select the robust dRLS-type algorithms to show their performance in terms of PSD at node~$1$. From the results, the proposed R-dRLS and DCD-R-dRLS algorithms have lower side lobes in the PSD curves than those of the other two algorithms, thus fitting much better the true spectrum.

\section{Conclusion}
In this paper, we have derived a new dRLS algorithm which is robust in impulsive noise, based on the minimization of a local RLS cost function with a time-dependent constraint on the squared norm of the intermediate estimate update. Following the diffusion strategy, the constraint is dynamically adjusted with the help of side information from the neighboring nodes. We also analyze the convergence of the proposed algorithm in the mean square sense under impulsive noise. Then, its DCD version was developed to reduce the computational complexity. Moreover, to adapt the proposed algorithms to an abrupt change of the unknown parameter vector, a non-stationary control approach has also been designed. Simulation results have verified that the proposed algorithms perform better than the known algorithms in impulsive noise scenarios.

\appendices
\numberwithin{equation}{section}
\section{Verification of \eqref{025}}
\begin{figure}[htb]
    \centering
    \includegraphics[scale=0.53] {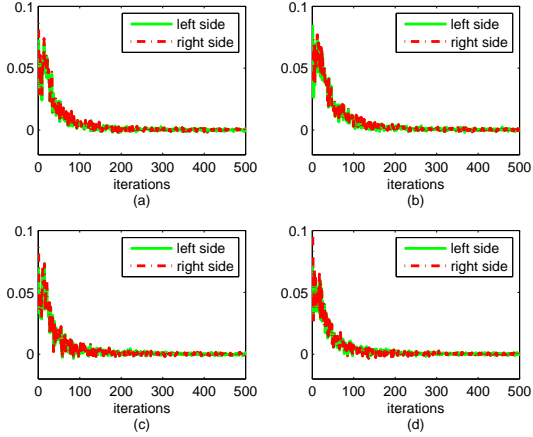}
    \hspace{2cm}\caption{Simulation results for \eqref{025} at different nodes in impulsive noise. (a) Node 1, (b) Node 6, (c) Node 11, and (d) Node 16 . Simulation setting is the same as for Fig. \ref{Fig5}.}
    \label{Fig16}
\end{figure}

\renewcommand{\theequation}{\thesection.\arabic{equation}}
From Fig.~\ref{Fig16}, one can see that the left side of \eqref{025} has a good agreement with the right side of that\footnote{ Similar results at other nodes have not been shown here because of the page limitation.}. This reveals that the simplification from \eqref{024} to \eqref{025} is reasonable.
\section{Convergence of $E\{\xi_k(i)\}$ to 0}
It is evident from \eqref{014} that $\xi_k(i)$ as a function of $i$ is non-increasing in adaptation process, with positive values. So, the limit of $E\{\xi_k(i)\}$ at $i \rightarrow \infty$ is existent. Applying the expectation operator to \eqref{014}, we obtain
\begin{equation}
\label{A38}
\begin{array}{rcl}
\begin{aligned}
E\left\lbrace \zeta_k(i) \right\rbrace & = \beta E \left\lbrace \xi_k(i-1) \right\rbrace + \\
&(1-\beta) E\left\lbrace \min [\lVert \bm g_{k,i}\rVert_2^2 e_k^2(i),\xi_k(i-1) ] \right\rbrace, \\
\end{aligned}
\end{array}
\end{equation}
\begin{equation}
\label{A39}
\begin{array}{rcl}
\begin{aligned}
E\left\lbrace \xi_k(i) \right\rbrace =&\sum \limits_{m\in\mathcal{N}_k}c_{m,k} E\left\lbrace \zeta_m(i)\right\rbrace ,
\end{aligned}
\end{array}
\end{equation}

Again using the assumption that the variance of $\xi_k(i)$ is small enough since $\beta$ closes to 1, we are able to make the approximation,
\begin{equation}
\begin{array}{rcl}
\begin{aligned}
\label{A40}
&E\left\lbrace \min [\lVert \bm g_{k,i}\rVert_2^2 e_k^2(i),\xi_k(i-1) ] \right\rbrace \approx \int \limits_{0}^{E\{\xi_k(i-1)\}}m_k dF_{k,i}(m_k)\\
&\;\;\;\;\;\;\;\;\;\;+E\{\xi_k(i-1)\}P_{k,i}[m_k>E\{\xi_k(i-1)\}]\\
\end{aligned}
\end{array}
\end{equation}
where $m_k\doteq \lVert \bm g_{k,i}\rVert_2^2 e_k^2(i)$ means that both $m_k$ and $\lVert \bm g_{k,i}\rVert_2^2 e_k^2(i)$ have the same distribution, $P_{k,i}[\cdot]$ denotes the probability of event in the argument, and $F_{k,i}(m_k)$ denotes the distribution function of $m_k$ at time instant $i$.

Let us define the network global vectors as follows:
\begin{equation}
\begin{array}{rcl}
\begin{aligned}
\label{A41}
\bm \xi(i)& \triangleq \text{col}\{\xi_1(i), ..., \xi_N(i)\}, \\
\bm \zeta(i)& \triangleq \text{col}\{\zeta_1(i), ..., \zeta_N(i)\}. \\
\end{aligned}
\end{array}
\end{equation}
Therefore, according to \eqref{A40} and \eqref{A41}, we reformulate \eqref{A38} and~\eqref{A39} for all nodes as:
\begin{equation}
\begin{array}{rcl}
\begin{aligned}
\label{A42}
E\{\bm \xi (i)\} =  & \bm C^T  \left[  \beta E\{\bm \xi(i-1)\} + \right. \\
&\left. (1-\beta) \left( \bm H_i E\{\bm \xi(i-1)\}+\bm m(i) \right) \right]
\end{aligned}
\end{array}
\end{equation}
where
\begin{equation}
\begin{array}{rcl}
\begin{aligned}
\label{A43}
\bm H_i = \text{diag}\left\lbrace P_{1,i}[m_1>E\{\xi_1(i-1)\}], ...,\right. \\
\left. P_{N,i}[m_N>E\{\xi_N(i-1)\}] \right\rbrace, \\
\end{aligned}
\end{array}
\end{equation}
and
\begin{equation}
\begin{array}{rcl}
\begin{aligned}
\label{A44}
\bm m_i = \text{col}\left\lbrace \int \limits_{0}^{E\{\xi_1(i-1)\}}m_1 dF_{1,i}(m_1), ...,\right. \\
\left. \int \limits_{0}^{E\{\xi_N(i-1)\}}m_N dF_{N,i}(m_N) \right\rbrace. \\
\end{aligned}
\end{array}
\end{equation}

Taking the $\infty$-norm of both sides of \eqref{A42} and recalling $\lVert \bm C^T \rVert_\infty=1$, it is found the following inequality:
\begin{equation}
\begin{array}{rcl}
\begin{aligned}
\label{A45}
\lVert E\{\bm \xi (i)\} \rVert_\infty \leq  &  \beta \lVert E\{\bm \xi(i-1)\} \rVert_\infty  +\\
&(1-\beta) \lVert \bm H_i E\{\bm \xi(i-1)\}+\bm m(i) \lVert_\infty.
\end{aligned}
\end{array}
\end{equation}

Based on the diagonal definition in \eqref{A43}, we deduce an equivalent form from \eqref{A45}, i.e., for $k=1,...,N$,
\begin{equation}
\begin{array}{rcl}
\begin{aligned}
\label{A46}
 E\{ \xi_k(i)\} &\leq \beta E\{ \xi_k(i-1)\}  + \\
(1-&\beta) E\{\xi_k(i-1)\}P_{k,i}[m_k>E\{\xi_k(i-1)\}] +\\
(1-&\beta) \int \limits_{0}^{E\{\xi_k(i-1)\}}m_k dF_{k,i}(m_k).
\end{aligned}
\end{array}
\end{equation}

It is supposed that there is a limit for $F_{k,i}(m_k)$ when $i \rightarrow \infty$, \eqref{A46} further reduces to
\begin{equation}
\begin{array}{rcl}
\begin{aligned}
\label{A47}
E\{\xi_k(\infty)\} \cdot P_{k,\infty}[m_k \leq E\{\xi_k(\infty)\}] \leq \\
\int_{0}^{E\{\xi_k(\infty)\}}m_k dF_{k,\infty}(m_k).
\end{aligned}
\end{array}
\end{equation}
In \eqref{A47}, the relation $ E\{ \xi_k(i)\}=E\{ \xi_k(i-1)\}$ as $i \rightarrow \infty$ is also used. Herein, we consider the equality case in \eqref{A47}, i.e.,
\begin{equation}
\begin{array}{rcl}
\begin{aligned}
\label{A48}
 E\{\xi_k(\infty)\} \cdot  P_{k,\infty}[m_k \leq E\{\xi_k(\infty)\}] = \\
\int_{0}^{E\{\xi_k(\infty)\}}m_k dF_{k,\infty}(m_k).
\end{aligned}
\end{array}
\end{equation}
It is shown in Appendix A in \cite{vega2008new}, for a similar
equation \eqref{A48}, its solution is $E\{\xi_k(\infty)\}]=0$. Since
\eqref{A48} is an upper bound of \eqref{A47}, we can conclude that
$E\{\xi_k(i)\}$ given by~\eqref{014} would also converge to zero.
Moreover, as its intermediate quantity, $E\{\zeta_k(i)\} $ also
converges to zero.

\ifCLASSOPTIONcaptionsoff
  \newpage
\fi

\bibliographystyle{IEEEtran}
\bibliography{IEEEabrv,mybibfile}

\begin{thebibliography}{10}
\providecommand{\url}[1]{#1}
\csname url@samestyle\endcsname
\providecommand{\newblock}{\relax}
\providecommand{\bibinfo}[2]{#2}
\providecommand{\BIBentrySTDinterwordspacing}{\spaceskip=0pt\relax}
\providecommand{\BIBentryALTinterwordstretchfactor}{4}
\providecommand{\BIBentryALTinterwordspacing}{\spaceskip=\fontdimen2\font plus
\BIBentryALTinterwordstretchfactor\fontdimen3\font minus
  \fontdimen4\font\relax}
\providecommand{\BIBforeignlanguage}[2]{{%
\expandafter\ifx\csname l@#1\endcsname\relax
\typeout{** WARNING: IEEEtran.bst: No hyphenation pattern has been}%
\typeout{** loaded for the language `#1'. Using the pattern for}%
\typeout{** the default language instead.}%
\else
\language=\csname l@#1\endcsname
\fi
#2}}
\providecommand{\BIBdecl}{\relax}
\BIBdecl

\bibitem{YYu2018}
Y.~Yu, H.~Zhao, R.~C. de~Lamare, and Y.~Zakharov, ``Robust diffusion recursive
  least squares estimation with side information for networked agents,'' in
  \emph{2018 IEEE International Conference on Acoustics, Speech and Signal
  Processing (ICASSP)}, April 2018, pp. 4099--4103.

\bibitem{sayed2014adaptive}
A.~H. Sayed, ``Adaptive networks,'' \emph{Proceedings of the IEEE}, vol. 102,
  no.~4, pp. 460--497, 2014.

\bibitem{sayed2014adaptation}
A.~H. Sayed, ``Adaptation, learning, and optimization over networks,''
  \emph{Foundations and Trends in Machine Learning}, vol.~7, no. 4-5, pp.
  311--801, 2014.

\bibitem{tu2011mobile}
S.-Y. Tu and A.~H. Sayed, ``Mobile adaptive networks,'' \emph{IEEE Journal of
  Selected Topics in Signal Processing}, vol.~5, no.~4, pp. 649--664, 2011.

\bibitem{sayed2013diffusion}
A.~H. Sayed, S.-Y. Tu, J.~Chen, X.~Zhao, and Z.~J. Towfic, ``Diffusion
  strategies for adaptation and learning over networks: an examination of
  distributed strategies and network behavior,'' \emph{IEEE Signal Processing
  Magazine}, vol.~30, no.~3, pp. 155--171, 2013.

\bibitem{kanna2015distributed}
S.~Kanna, D.~H. Dini, Y.~Xia, S.~Hui, and D.~P. Mandic, ``Distributed widely
  linear kalman filtering for frequency estimation in power networks,''
  \emph{IEEE Transactions on Signal and Information Processing over Networks},
  vol.~1, no.~1, pp. 45--57, 2015.

\bibitem{di2013distributed}
P.~Di~Lorenzo, S.~Barbarossa, and A.~H. Sayed, ``Distributed spectrum
  estimation for small cell networks based on sparse diffusion adaptation,''
  \emph{IEEE Signal Processing Letters}, vol.~20, no.~12, pp. 1261--1265, 2013.

\bibitem{miller2016distributed}
T.~G. Miller, S.~Xu, R.~C. de~Lamare, and H.~V. Poor, ``Distributed spectrum
  estimation based on alternating mixed discrete-continuous adaptation,''
  \emph{IEEE Signal Processing Letters}, vol.~23, no.~4, pp. 551--555, 2016.

\bibitem{li2010distributed}
L.~Li, J.~A. Chambers, C.~G. Lopes, and A.~H. Sayed, ``Distributed estimation
  over an adaptive incremental network based on the affine projection
  algorithm,'' \emph{IEEE Transactions on Signal Processing}, vol.~58, no.~1,
  pp. 151--164, 2010.

\bibitem{tu2012diffusion}
S.~Y. Tu and A.~H. Sayed, ``Diffusion strategies outperform consensus
  strategies for distributed estimation over adaptive networks,'' \emph{IEEE
  Transactions on Signal Processing}, vol.~60, no.~12, pp. 6217--6234, 2012.

\bibitem{lopes2008diffusion}
C.~G. Lopes and A.~H. Sayed, ``Diffusion least-mean squares over adaptive
  networks: Formulation and performance analysis,'' \emph{IEEE Transactions on
  Signal Processing}, vol.~56, no.~7, pp. 3122--3136, 2008.

\bibitem{chen2012diffusion}
J.~Chen and A.~H. Sayed, ``Diffusion adaptation strategies for distributed
  optimization and learning over networks,'' \emph{IEEE Transactions on Signal
  Processing}, vol.~60, no.~8, pp. 4289--4305, 2012.

\bibitem{xu2015adaptive}
S.~Xu, R.~C. de~Lamare, and H.~V. Poor, ``Adaptive link selection algorithms
  for distributed estimation,'' \emph{EURASIP Journal on Advances in Signal
  Processing}, vol. 2015, no.~1, p.~86, 2015.

\bibitem{LU2018243}
L.~Lu, H.~Zhao, and B.~Champagne, ``Diffusion total least-squares algorithm
  with multi-node feedback,'' \emph{Signal Processing}, vol. 153, pp. 243--254,
  2018.

\bibitem{lee2015variable}
H.-S. Lee, S.-E. Kim, J.-W. Lee, and W.-J. Song, ``A variable step-size
  diffusion {LMS} algorithm for distributed estimation.'' \emph{IEEE Trans.
  Signal Processing}, vol.~63, no.~7, pp. 1808--1820, 2015.

\bibitem{han2017non}
H.~Han, S.~Zhang, and H.~Liang, ``Non-parametric variable step-size diffusion
  {LMS} algorithm over adaptive networks,'' \emph{Electronics Letters}, 2017.

\bibitem{aifir}
R.~C. de~Lamare and R.~Sampaio-Neto, ``Adaptive reduced-rank mmse filtering
  with interpolated fir filters and adaptive interpolators,'' \emph{IEEE Signal
  Processing Letters}, vol.~12, no.~3, pp. 177--180, March 2005.

\bibitem{miller1976detection}
J.~Miller and J.~Thomas, ``The detection of signals in impulsive noise modeled
  as a mixture process,'' \emph{IEEE Transactions on Communications}, vol.~24,
  no.~5, pp. 559--563, 1976.

\bibitem{blackard1993measurements}
K.~L. Blackard, T.~S. Rappaport, and C.~W. Bostian, ``Measurements and models
  of radio frequency impulsive noise for indoor wireless communications,''
  \emph{IEEE Journal on Selected Areas in Communications}, vol.~11, no.~7, pp.
  991--1001, 1993.

\bibitem{jio}
R.~C. de~Lamare and R.~Sampaio-Neto, ``Reduced-rank adaptive filtering based on
  joint iterative optimization of adaptive filters,'' \emph{IEEE Signal
  Processing Letters}, vol.~14, no.~12, pp. 980--983, Dec 2007.

\bibitem{rrsgp}
M.~Yukawa, R.~C. de~Lamare, and I.~Yamada, ``Robust reduced-rank adaptive
  algorithm based on parallel subgradient projection and krylov subspace,''
  \emph{IEEE Transactions on Signal Processing}, vol.~57, no.~12, pp.
  4660--4674, Dec 2009.

\bibitem{spa}
R.~C.~D. Lamare and R.~Sampaio-Neto, ``Minimum mean-squared error iterative
  successive parallel arbitrated decision feedback detectors for ds-cdma
  systems,'' \emph{IEEE Transactions on Communications}, vol.~56, no.~5, pp.
  778--789, May 2008.

\bibitem{zoubir2012robust}
A.~M. Zoubir, V.~Koivunen, Y.~Chakhchoukh, and M.~Muma, ``Robust estimation in
  signal processing: A tutorial-style treatment of fundamental concepts,''
  \emph{IEEE Signal Processing Magazine}, vol.~29, no.~4, pp. 61--80, 2012.

\bibitem{smtvb}
R.~C. de~Lamare and P.~S.~R. Diniz, ``Set-membership adaptive algorithms based
  on time-varying error bounds for cdma interference suppression,'' \emph{IEEE
  Transactions on Vehicular Technology}, vol.~58, no.~2, pp. 644--654, Feb
  2009.

\bibitem{jidf}
R.~C. de~Lamare and R.~Sampaio-Neto, ``Adaptive reduced-rank processing based
  on joint and iterative interpolation, decimation, and filtering,'' \emph{IEEE
  Transactions on Signal Processing}, vol.~57, no.~7, pp. 2503--2514, July
  2009.

\bibitem{jidf_echo}
M.~Yukawa, R.~C. de~Lamare, and R.~Sampaio-Neto, ``Efficient acoustic echo
  cancellation with reduced-rank adaptive filtering based on selective
  decimation and adaptive interpolation,'' \emph{IEEE Transactions on Audio,
  Speech, and Language Processing}, vol.~16, no.~4, pp. 696--710, May 2008.

\bibitem{sjidf}
R.~Fa, R.~C. de~Lamare, and L.~Wang, ``Reduced-rank stap schemes for airborne
  radar based on switched joint interpolation, decimation and filtering
  algorithm,'' \emph{IEEE Transactions on Signal Processing}, vol.~58, no.~8,
  pp. 4182--4194, Aug 2010.

\bibitem{ccg}
L.~Wang and R.~C.~D. Lamare, ``Constrained adaptive filtering algorithms based
  on conjugate gradient techniques for beamforming,'' \emph{IET Signal
  Processing}, vol.~4, no.~6, pp. 686--697, Dec 2010.

\bibitem{jiocdma}
R.~C. de~Lamare and R.~Sampaio-Neto, ``Reduced-rank space–time adaptive
  interference suppression with joint iterative least squares algorithms for
  spread-spectrum systems,'' \emph{IEEE Transactions on Vehicular Technology},
  vol.~59, no.~3, pp. 1217--1228, March 2010.

\bibitem{jiomimo}
R.~C. de~Lamare and R.~Sampaio-Neto, ``Adaptive reduced-rank equalization
  algorithms based on alternating optimization design techniques for mimo
  systems,'' \emph{IEEE Transactions on Vehicular Technology}, vol.~60, no.~6,
  pp. 2482--2494, July 2011.

\bibitem{tds}
P.~Clarke and R.~C. de~Lamare, ``Transmit diversity and relay selection
  algorithms for multirelay cooperative mimo systems,'' \emph{IEEE Transactions
  on Vehicular Technology}, vol.~61, no.~3, pp. 1084--1098, March 2012.

\bibitem{mbdf}
R.~C. de~Lamare, ``Adaptive and iterative multi-branch mmse decision feedback
  detection algorithms for multi-antenna systems,'' \emph{IEEE Transactions on
  Wireless Communications}, vol.~12, no.~10, pp. 5294--5308, October 2013.

\bibitem{rrstap}
R.~Fa and R.~C.~D. Lamare, ``Reduced-rank stap algorithms using joint iterative
  optimization of filters,'' \emph{IEEE Transactions on Aerospace and
  Electronic Systems}, vol.~47, no.~3, pp. 1668--1684, July 2011.

\bibitem{l1stap}
Z.~Yang, R.~C.~D. Lamare, and X.~Li, ``Sparsity-aware space-time adaptive
  processing algorithms with l1-norm regularisation for airborne radar,''
  \emph{IET Signal Processing}, vol.~6, no.~5, pp. 413--423, July 2012.

\bibitem{l1stap2}
Z.~Yang, R.~C. de~Lamare, and X.~Li, ``$l_1$ -regularized stap algorithms with
  a generalized sidelobe canceler architecture for airborne radar,'' \emph{IEEE
  Transactions on Signal Processing}, vol.~60, no.~2, pp. 674--686, Feb 2012.

\bibitem{rccm}
L.~Landau, R.~C. de~Lamare, and M.~Haardt, ``Robust adaptive beamforming
  algorithms using the constrained constant modulus criterion,'' \emph{IET
  Signal Processing}, vol.~8, no.~5, pp. 447--457, July 2014.

\bibitem{dfjio}
L.~Wang, R.~C. de~Lamare, and M.~Haardt, ``Direction finding algorithms based
  on joint iterative subspace optimization,'' \emph{IEEE Transactions on
  Aerospace and Electronic Systems}, vol.~50, no.~4, pp. 2541--2553, October
  2014.

\bibitem{locsme}
H.~Ruan and R.~C. de~Lamare, ``Robust adaptive beamforming using a
  low-complexity shrinkage-based mismatch estimation algorithm,'' \emph{IEEE
  Signal Processing Letters}, vol.~21, no.~1, pp. 60--64, Jan 2014.

\bibitem{rrser}
Y.~Cai, R.~C. de~Lamare, B.~Champagne, B.~Qin, and M.~Zhao, ``Adaptive
  reduced-rank receive processing based on minimum symbol-error-rate criterion
  for large-scale multiple-antenna systems,'' \emph{IEEE Transactions on
  Communications}, vol.~63, no.~11, pp. 4185--4201, Nov 2015.

\bibitem{rdrcb}
S.~D. Somasundaram, N.~H. Parsons, P.~Li, and R.~C. de~Lamare,
  ``Reduced-dimension robust capon beamforming using krylov-subspace
  techniques,'' \emph{IEEE Transactions on Aerospace and Electronic Systems},
  vol.~51, no.~1, pp. 270--289, January 2015.

\bibitem{rrdoa}
L.~Qiu, Y.~Cai, R.~C. de~Lamare, and M.~Zhao, ``Reduced-rank doa estimation
  algorithms based on alternating low-rank decomposition,'' \emph{IEEE Signal
  Processing Letters}, vol.~23, no.~5, pp. 565--569, May 2016.

\bibitem{okspme}
H.~Ruan and R.~C. de~Lamare, ``Robust adaptive beamforming based on low-rank
  and cross-correlation techniques,'' \emph{IEEE Transactions on Signal
  Processing}, vol.~64, no.~15, pp. 3919--3932, Aug 2016.

\bibitem{kaesprit}
S.~F.~B. Pinto and R.~C. de~Lamare, ``Multistep knowledge-aided iterative
  esprit: Design and analysis,'' \emph{IEEE Transactions on Aerospace and
  Electronic Systems}, vol.~54, no.~5, pp. 2189--2201, Oct 2018.

\bibitem{georgiou1999alpha}
P.~G. Georgiou, P.~Tsakalides, and C.~Kyriakakis, ``Alpha-stable modeling of
  noise and robust time-delay estimation in the presence of impulsive noise,''
  \emph{IEEE transactions on Multimedia}, vol.~1, no.~3, pp. 291--301, 1999.

\bibitem{vega2008new}
L.~R. Vega, H.~Rey, J.~Benesty, and S.~Tressens, ``A new robust variable
  step-size {NLMS} algorithm,'' \emph{IEEE Transactions on Signal Processing},
  vol.~56, no.~5, pp. 1878--1893, 2008.

\bibitem{Chitre2006}
M.~A. Chitre, J.~R. Potter, and S.~Ong, ``Optimal and near-optimal signal
  detection in snapping shrimp dominated ambient noise,'' \emph{IEEE Journal of
  Oceanic Engineering}, vol.~31, no.~2, pp. 497--503, 2006.

\bibitem{Bouvet1989compa}
M.~Bouvet and S.~C. Schwartz, ``Comparison of adaptive and robust receivers for
  signal detection in ambient underwater noise,'' \emph{IEEE Transactions on
  Acoustics, Speech, and Signal Processing}, vol.~37, no.~5, pp. 621--626,
  1989.

\bibitem{ZHU201594}
X.~Zhu, W.-P. Zhu, and B.~Champagne, ``Spectrum sensing based on fractional
  lower order moments for cognitive radios in $\alpha$-stable distributed
  noise,'' \emph{Signal Processing}, vol. 111, pp. 94--105, 2015.

\bibitem{al2017robust}
S.~Al-Sayed, A.~M. Zoubir, and A.~H. Sayed, ``Robust distributed estimation by
  networked agents,'' \emph{IEEE Transactions on Signal Processing}, vol.~65,
  no.~15, pp. 3909--3921, Aug. 2017.

\bibitem{wen2013diffusion}
F.~Wen, ``Diffusion least-mean p-power algorithms for distributed estimation in
  alpha-stable noise environments,'' \emph{Electronics letters}, vol.~49,
  no.~21, pp. 1355--1356, 2013.

\bibitem{ni2016diffusion}
J.~Ni, J.~Chen, and X.~Chen, ``Diffusion sign-error {LMS} algorithm:
  formulation and stochastic behavior analysis,'' \emph{Signal Processing},
  vol. 128, pp. 142--149, 2016.

\bibitem{ma2016diffusion}
W.~Ma, B.~Chen, J.~Duan, and H.~Zhao, ``Diffusion maximum correntropy criterion
  algorithms for robust distributed estimation,'' \emph{Digital Signal
  Processing}, vol.~58, pp. 10--19, 2016.

\bibitem{8405555}
Y.~He, F.~Wang, S.~Wang, P.~Ren, and B.~Chen, ``Maximum total correntropy
  diffusion adaptation over networks with noisy links,'' \emph{IEEE
  Transactions on Circuits and Systems II: Express Briefs}, 2018.

\bibitem{chouvardas2011adaptive}
S.~Chouvardas, K.~Slavakis, and S.~Theodoridis, ``Adaptive robust distributed
  learning in diffusion sensor networks,'' \emph{IEEE Transactions on Signal
  Processing}, vol.~59, no.~10, pp. 4692--4707, 2011.

\bibitem{cattivelli2008diffusion}
F.~S. Cattivelli, C.~G. Lopes, and A.~H. Sayed, ``Diffusion recursive
  least-squares for distributed estimation over adaptive networks,'' \emph{IEEE
  Transactions on Signal Processing}, vol.~56, no.~5, pp. 1865--1877, 2008.

\bibitem{vahidpour2017analysis}
V.~Vahidpour, A.~Rastegarnia, A.~Khalili, and S.~Sanei, ``Analysis of partial
  diffusion recursive least squares adaptation over noisy links,'' \emph{IET
  Signal Processing}, vol.~11, no.~6, pp. 749--758, 2017.

\bibitem{mateos2009distributed}
G.~Mateos, I.~D. Schizas, and G.~B. Giannakis, ``Distributed recursive
  least-squares for consensus-based in-network adaptive estimation,''
  \emph{IEEE Transactions on Signal Processing}, vol.~57, no.~11, pp.
  4583--4588, 2009.

\bibitem{wang2018decentralized}
Z.~Wang, Z.~Yu, Q.~Ling, D.~Berberidis, and G.~B. Giannakis, ``Decentralized
  {RLS} with data-adaptive censoring for regressions over large-scale
  networks,'' \emph{IEEE Transactions on Signal Processing}, vol.~66, no.~6,
  pp. 1634--1648, 2018.

\bibitem{vega2009fast}
L.~R. Vega, H.~Rey, J.~Benesty, and S.~Tressens, ``A fast robust recursive
  least-squares algorithm,'' \emph{IEEE Transactions on Signal Processing},
  vol.~57, no.~3, pp. 1209--1216, 2009.

\bibitem{bhotto2011robust}
M.~Z.~A. Bhotto and A.~Antoniou, ``Robust recursive least-squares
  adaptive-filtering algorithm for impulsive-noise environments,'' \emph{IEEE
  Signal processing letters}, vol.~18, no.~3, pp. 185--188, 2011.

\bibitem{ma2016robust}
W.~Ma, J.~Duan, G.~Gui, and B.~Chen, ``Robust diffusion recursive adaptive
  filtering algorithm based on lp-norm,'' in \emph{2016 35th Chinese Control
  Conference (CCC)}.\hskip 1em plus 0.5em minus 0.4em\relax IEEE, 2016, pp.
  1404--1408.

\bibitem{zakharov2008low}
Y.~Zakharov, G.~P. White, and J.~Liu, ``Low-complexity {RLS} algorithms using
  dichotomous coordinate descent iterations,'' \emph{IEEE Transactions on
  Signal Processing}, vol.~56, no.~7, pp. 3150--3161, 2008.

\bibitem{zakharov2004multiplication}
Y.~Zakharov and T.~Tozer, ``Multiplication-free iterative algorithm for {LS}
  problem,'' \emph{Electronics Letters}, vol.~40, no.~9, pp. 567--569, 2004.

\bibitem{liu2009architecture}
J.~Liu, Y.~Zakharov, and B.~Weaver, ``Architecture and {FPGA} design of
  dichotomous coordinate descent algorithms,'' \emph{IEEE Transactions on
  Circuits and Systems I: Regular Papers}, vol.~56, no.~11, pp. 2425--2438,
  2009.

\bibitem{zakharov2013dcd}
Y.~V. Zakharov and V.~H. Nascimento, ``{DCD-RLS} adaptive filters with
  penalties for sparse identification,'' \emph{IEEE transactions on signal
  processing}, vol.~61, no.~12, pp. 3198--3213, 2013.

\bibitem{arablouei2013reduced}
R.~Arablouei, K.~Do{\u{g}}an{\c{c}}ay, and S.~Werner, ``Reduced-complexity
  distributed least-squares estimation over adaptive networks,'' in \emph{2013
  IEEE 14th Workshop on Signal Processing Advances in Wireless Communications
  (SPAWC)}.\hskip 1em plus 0.5em minus 0.4em\relax IEEE, 2013, pp. 150--154.

\bibitem{sayed2011adaptive}
A.~H. Sayed, \emph{Adaptive filters}.\hskip 1em plus 0.5em minus 0.4em\relax
  John Wiley \& Sons, 2011.

\bibitem{bershad2008error}
N.~J. Bershad, ``On error saturation nonlinearities for {LMS} adaptation in
  impulsive noise,'' \emph{IEEE Transactions on Signal Processing}, vol.~56,
  no.~9, pp. 4526--4529, 2008.

\bibitem{shao1993signal}
M.~Shao and C.~Nikias, ``Signal processing with fractional lower order moments:
  stable processes and their applications,'' \emph{Proceedings of the IEEE},
  vol.~81, no.~7, pp. 986--1010, 1993.

\bibitem{CHEN2018318}
B.~Chen, X.~Wang, N.~Lu, S.~Wang, J.~Cao, and J.~Qin, ``Mixture correntropy for
  robust learning,'' \emph{Pattern Recognition}, vol.~79, pp. 318--327, 2018.

\bibitem{amari1998natural}
S.-I. Amari and S.~C. Douglas, ``Why natural gradient?'' in \emph{Proceedings
  of the IEEE international conference on Acoustics, Speech and Signal
  Processing}, vol.~2, 1998, pp. 1213--1216.

\bibitem{pelekanakis2014adaptive}
K.~Pelekanakis and M.~Chitre, ``Adaptive sparse channel estimation under
  symmetric alpha-stable noise,'' \emph{IEEE Transactions on Wireless
  Communications}, vol.~13, no.~6, pp. 3183--3195, 2014.

\bibitem{takahashi2010diffusion}
N.~Takahashi, I.~Yamada, and A.~H. Sayed, ``Diffusion least-mean squares with
  adaptive combiners: Formulation and performance analysis,'' \emph{IEEE
  Transactions on Signal Processing}, vol.~58, no.~9, pp. 4795--4810, 2010.

\bibitem{6854838}
J.~Fernandez-Bes, J.~Arenas-García, and A.~H. Sayed, ``Adjustment of
  combination weights over adaptive diffusion networks,'' in \emph{2014 IEEE
  International Conference on Acoustics, Speech and Signal Processing
  (ICASSP)}, May. 2014, pp. 6409--6413.

\bibitem{7936509}
R.~Abdolee and V.~Vakilian, ``An iterative scheme for computing combination
  weights in diffusion wireless networks,'' \emph{IEEE Wireless Communications
  Letters}, vol.~6, no.~4, pp. 510--513, 2017.

\bibitem{ahn2017new}
D.-C. Ahn, J.-W. Lee, S.-J. Shin, and W.-J. Song, ``A new robust variable
  weighting coefficients diffusion {LMS} algorithm,'' \emph{Signal Processing},
  vol. 131, pp. 300--306, 2017.

\bibitem{song2013normalized}
I.~Song, P.~Park, and R.~W. Newcomb, ``A normalized least mean squares
  algorithm with a step-size scaler against impulsive measurement noise,''
  \emph{IEEE Transactions on Circuits and Systems II: Express Briefs}, vol.~60,
  no.~7, pp. 442--445, 2013.

\bibitem{hur2017variable}
J.~Hur, I.~Song, and P.~Park, ``A variable step-size normalized subband
  adaptive filter with a step-size scaler against impulsive measurement
  noise,'' \emph{IEEE Transactions on Circuits and Systems II: Express Briefs},
  vol.~64, no.~7, pp. 842--846, 2017.

\bibitem{boyd2004convex}
S.~Boyd and L.~Vandenberghe, \emph{Convex optimization}.\hskip 1em plus 0.5em
  minus 0.4em\relax Cambridge university press, 2004.

\bibitem{chen2013distributed}
J.~Chen and A.~H. Sayed, ``Distributed pareto optimization via diffusion
  strategies,'' \emph{IEEE Journal of Selected Topics in Signal Processing},
  vol.~7, no.~2, pp. 205--220, 2013.

\bibitem{zhang2017variable}
L.~Zhang, Y.~Cai, C.~Li, and R.~C. de~Lamare, ``Variable forgetting factor
  mechanisms for diffusion recursive least squares algorithm in sensor
  networks,'' \emph{EURASIP Journal on Advances in Signal Processing}, vol.
  2017, no.~1, p.~57, 2017.

\bibitem{zhou2011new}
Y.~Zhou, S.~Chan, and K.~Ho, ``New sequential partial-update least mean
  {M}-estimate algorithms for robust adaptive system identification in
  impulsive noise,'' \emph{IEEE Transactions on Industrial Electronics},
  vol.~58, no.~9, pp. 4455--4470, 2011.

\bibitem{al2003transient}
T.~Y. Al-Naffouri and A.~H. Sayed, ``Transient analysis of adaptive filters
  with error nonlinearities,'' \emph{IEEE Transactions on Signal Processing},
  vol.~51, no.~3, pp. 653--663, 2003.

\bibitem{price1958useful}
R.~Price, ``A useful theorem for nonlinear devices having gaussian inputs,''
  \emph{IRE Transactions on Information Theory}, vol.~4, no.~2, pp. 69--72,
  1958.

\bibitem{dogancay2008partial}
K.~Dogancay, \emph{Partial-update adaptive signal processing: Design Analysis
  and Implementation}.\hskip 1em plus 0.5em minus 0.4em\relax Academic Press,
  2008.

\end{thebibliography}





%






\end{document}